\pdfoutput=1
\documentclass[10pt,twocolumn,letterpaper]{article}

\usepackage{cvpr} %

\usepackage{colortbl}
\usepackage{arydshln}
\usepackage{multirow}
\definecolor{metrictint}{HTML}{2196F3}
\newcommand{\mc}[2]{\cellcolor{metrictint!#1}#2}

\makeatletter
\providecommand\captionof[1]{\def\@captype{#1}\caption}
\makeatother
\usepackage{tikz}

\definecolor{cvprblue}{rgb}{0.21,0.49,0.74}
\usepackage[pagebackref,breaklinks,colorlinks,allcolors=cvprblue]{hyperref}

\def\confName{CVPR}
\def\confYear{2027}

\title{\textsc{NavRSI}\,\raisebox{-0.2\height}{\includegraphics[height=1.9em]{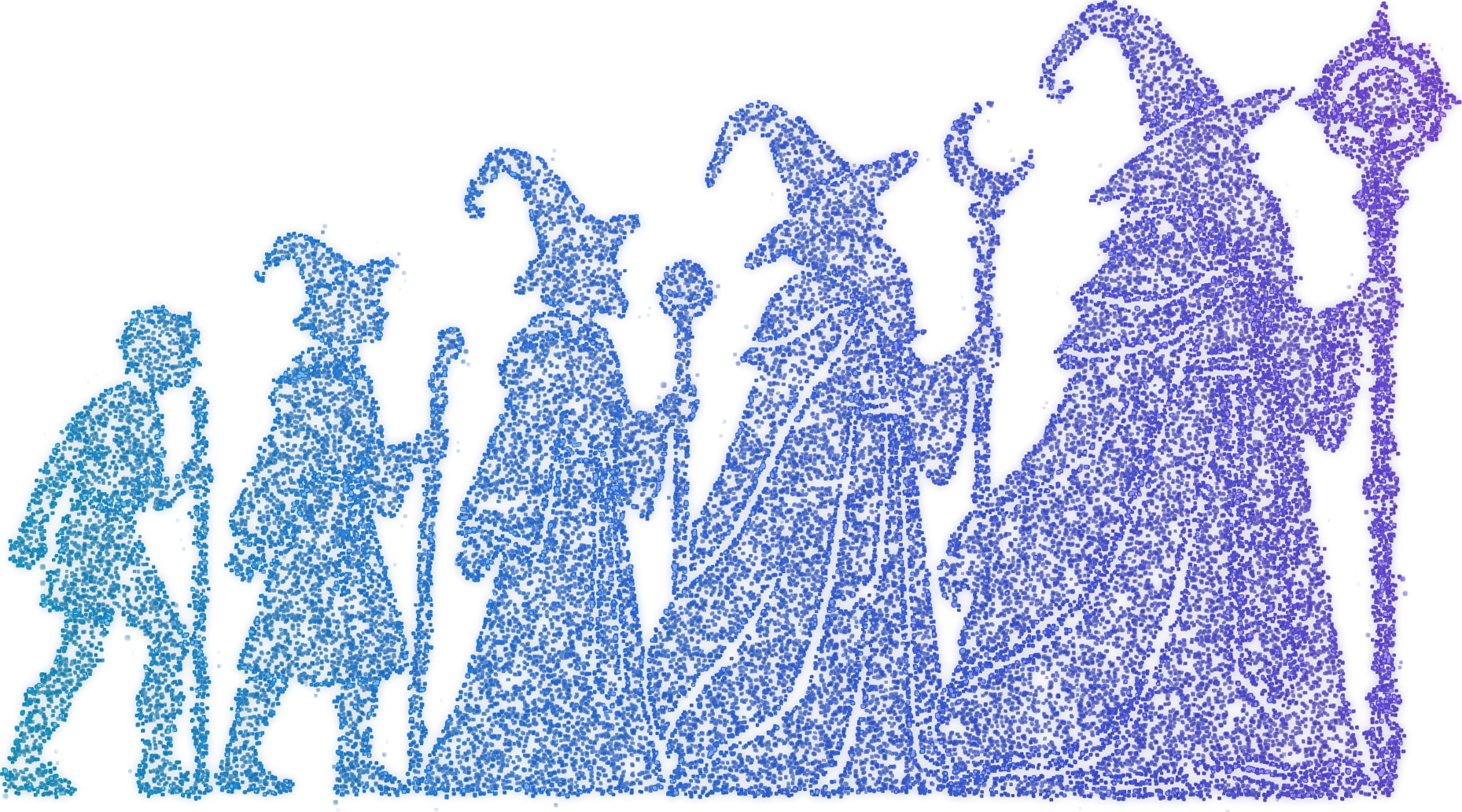}}: Recursive Harness Self-Improvement for \mbox{Vision-and-Language Navigation}}

\author{Guangzhao Dai$^{1}$, Guoxin Lian$^{2}$, Shuo Wang$^{2}$, Shengbin Yue$^{3}$,\\
Qianru Sun$^{1}$, Xiangbo Shu$^{4}$, Bin Zhu$^{1\dagger}$\\
$^{1}$School of Computing and Information Systems, Singapore Management University\\
$^{2}$School of Information, Renmin University of China\\
$^{3}$School of Data Science, Fudan University\\
$^{4}$School of Computer Science and Engineering, Nanjing University of Science and Technology\\
{\small $^{\dagger}$Project leader and corresponding author}\\[3pt]
{Project website: \href{https://daiguangzhao.github.io/NavRSI}{https://daiguangzhao.github.io/NavRSI/}}
}

\begin{document}
\twocolumn[{%
\renewcommand\twocolumn[1][]{#1}%
\maketitle
\begin{center}
\includegraphics[width=\textwidth]{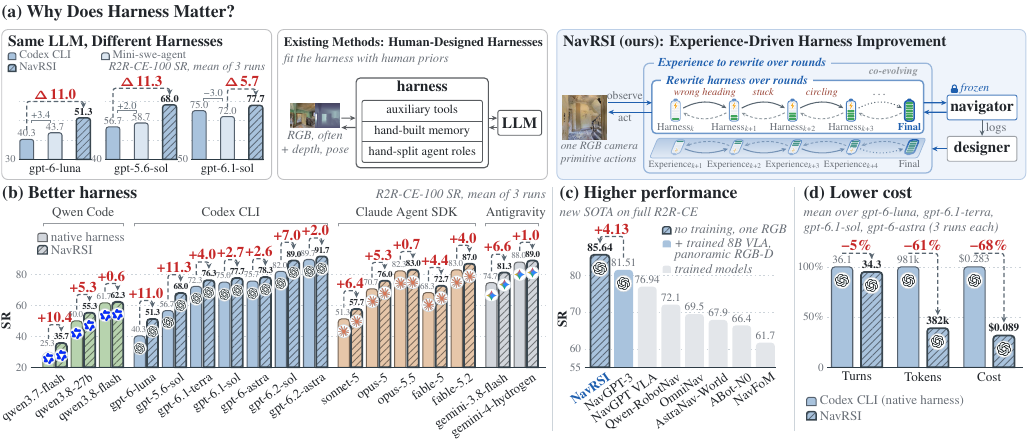}
\captionof{figure}{%
(a) The same LLM performs very differently under different harnesses. While existing harness-based VLN methods rely on human-designed mechanisms, NavRSI improves the harness through closed-loop navigation experience without modifying the LLM.
(b) NavRSI consistently improves navigation success rate over native harnesses across four different LLM families on R2R-CE-100.
(c) With GPT-6-Astra, NavRSI achieves \textbf{85.64}\% SR on the full R2R-CE val-unseen split, outperforming NavGPT-3 by \textbf{4.13}\%.
(d) Compared with Codex CLI, NavRSI reduces tool calls, token consumption, and API cost by \textbf{5}\%, \textbf{61}\%, and \textbf{68}\% on average over four GPT models.}
\label{fig:teaser}
\end{center}

}]
\begin{abstract}
Frontier large language models (LLMs) have enabled vision-and-language navigation (VLN) through direct closed-loop perception and action. However, navigation performance depends strongly on the harness governing LLM–environment interactions, while existing harnesses largely rely on manually designed navigation tools, memory, and agent roles. We introduce \textbf{NavRSI}, a harness-centric recursive self-improvement framework that optimizes the navigation behavior of frozen LLMs through closed-loop experience. In each round, NavRSI iteratively collects navigation trajectories and failure diagnostics, employs a designer agent to rewrite the harness, and retains the revisions that prove effective, while the designer accumulates experience from previous revisions so that the harness and this experience co-evolve. Privileged simulator diagnostics guide only the designer and remain inaccessible to the navigating LLM, enabling deployment with a single RGB camera and primitive actions, without task-specific model training or additional navigation tools. Experiments demonstrate consistent improvements across LLMs from four vendors, with \textbf{85.64}\% success rate on the full R2R-CE val-unseen split with GPT-6-Astra, surpassing the previously reported best by \textbf{4.13}\%. Evaluations on RxR-CE, NavRAG-CE, and a real humanoid robot further demonstrate its generalizability. \textit{Source code will be released.}
\end{abstract}

\section{Introduction}
\label{sec:intro}

Vision-and-language navigation (VLN) requires a robot to follow language instructions and navigate to the target destination in an unseen environment~\cite{anderson2018vision,krantz2020beyond,anderson2020rxr}. Existing approaches either train specialized navigation models on large-scale embodied data~\cite{hong2022bridging,an2025etpnav,zhang2024navid,cheng2025navila,wei2026streamvln,zhang2025embodied,qwennav} or construct explicit navigation workflows around general-purpose large language models (LLMs)~\cite{zhou2024navgpt,chen2024mapgpt,long2024instructnav,qiao2025opennav,shi2025smartway}. Frontier LLMs such as GPT-6-Astra and Fable-5.2 have recently opened up a promising alternative, where the LLM directly perceives visual observations and executes primitive actions through closed-loop interaction~\cite{zhou2026embodied}. This is enabled by a harness (such as Codex CLI and Claude Agent SDK), \textit{i.e.,} the program that connects the LLM to the environment and decides its context, tools, and feedback~\cite{zhong2026ai,zhou2026navgpt3}. The design of this harness thus becomes an important factor in translating the capabilities of LLMs into successful navigation~\cite{li2026agenticnav,chen2026harnessvln,zhao2026navharness,shi2026navharness}.

However, existing harness-based VLN methods~\cite{zhou2026embodied,li2026agenticnav,chen2026harnessvln,zhao2026navharness,shi2026navharness,zhou2026navgpt3,sheng2026ask} largely rely on human-designed assumptions about what an LLM needs to navigate successfully. They introduce auxiliary tools for depth estimation, waypoint execution, or learned navigation policies~\cite{li2026agenticnav,chen2026harnessvln,zhou2026navgpt3,cheng2026asena}, maintain explicit spatial or episodic memories~\cite{chen2026harnessvln,zhao2026navharness,sheng2026ask}, or divide navigation into specialized agent roles~\cite{shi2026navharness,chen2026harnessvln,zhou2026navgpt3}. Although effective, these designs prescribe how the LLM should perceive, remember, and act, rather than identifying its limitations from actual interaction with the environment. Importantly, the same LLM performs very differently under different harnesses. As shown in \cref{fig:teaser}(a), GPT-5.6-Sol achieves success rates of 56.7\% and 58.7\% with Codex CLI and Mini-swe-agent, but reaches 68.0\% with our NavRSI harness on the same primitive actions. The harness itself can thus be a significant bottleneck, raising a fundamental question: \textit{\textbf{Can we systematically improve the harness based on an LLM's own closed-loop behavior, rather than relying on manually designed navigation mechanisms?}}

Answering this question is challenging because navigation failures emerge from long-horizon interactions between perception, reasoning, and action. An agent may repeatedly attempt blocked movements, lose track of its heading, revisit explored locations, or stop before reaching the destination. Such failures are difficult to anticipate through static prompt design, and an overall success score gives limited guidance on how the harness should be revised. Moreover, the simulator diagnostics that explain these failures are unavailable to the deployed robot, so the optimization must produce a harness that operates without privileged information. Recursive self-improvement (RSI) has recently shown promise for optimizing agent code and execution scaffolds~\cite{zelikman2023self,zhang2025darwin,lee2026meta}, yet applying it to embodied navigation must account for the sequential, partially observable, and physically grounded nature of this interaction.

In this paper, we introduce \textbf{NavRSI}, \textbf{\textit{a harness-centric RSI framework}} that keeps the LLM frozen and iteratively rewrites the executable harness around it from closed-loop navigation experience, instead of modifying the model or adding navigation tools. As shown in \cref{fig:overview}, NavRSI starts from a general-purpose Mini-swe-agent harness~\cite{zhou2026embodied} and alternates between three steps: (1) \textit{running navigation episodes to collect closed-loop trajectories and failure diagnostics}, (2) \textit{employing a designer agent to analyze them and generate revised harness candidates}, and (3) \textit{evaluating the candidates to retain effective improvements.} The designer also maintains a record of its optimization experience, including observed failure patterns, the hypotheses behind its revisions, and evidence of their effectiveness, so that the harness and this experience co-evolve across rounds. To bridge the gap between optimization and deployment, the designer, a separate LLM, reads execution logs and simulator diagnostics (\textit{e.g.,} distance to the goal) from training episodes, whereas the frozen LLM, as the navigator, never sees these privileged signals and relies solely on RGB observations and action feedback. Through this process, NavRSI discovers a lightweight harness that provides action feedback, progress tracking, and arrival verification, without requiring task-specific model training, additional navigation tools, or specialized agent roles.

Extensive experiments demonstrate the effectiveness of NavRSI. On R2R-CE-100, the NavRSI harness consistently improves navigation success rate across four LLM families, outperforming their native harnesses by up to 11.3 SR (\cref{fig:teaser}(b)). With GPT-6-Astra, NavRSI sets new SOTA on the full R2R-CE val-unseen without training (85.64 SR), \textbf{+4.13} SR over \mbox{NavGPT-3}, which pairs the same LLM with an 8B model additionally trained on 19.28M samples (\cref{fig:teaser}(c), \cref{tab:r2r_full}). Evaluations on \mbox{RxR-CE}, \mbox{NavRAG-CE}~\cite{wang2025navrag}, and a Unitree G1 humanoid robot further demonstrate cross-dataset and sim-to-real transfer. Ablations show that its components are complementary and that closed-loop RSI outperforms one-shot writing.

Our main contributions are summarized as follows:
\begin{itemize}
\item We investigate the harness as an optimization target and introduce \textbf{NavRSI}, which improves the runtime interaction of frozen LLMs from their closed-loop navigation behavior, without task-specific model training.

\item We develop an iterative optimization framework in which a designer agent analyzes navigation failures, rewrites executable harness code, and accumulates experience from previous revisions. Simulator diagnostics guide only the designer and are never visible to the navigator, which bridges optimization and deployment.

\item NavRSI discovers a lightweight harness that operates with a single RGB camera and primitive actions, achieving strong performance across multiple LLM families, VLN benchmarks, and real-world environments, with competitive runtime efficiency.

\end{itemize}

\section{Related work}
\label{sec:related}

\noindent\textbf{Vision-and-language navigation.}
VLN was first studied on discrete navigation graphs~\cite{anderson2018vision} and later extended to continuous
environments with low-level actions~\cite{krantz2020beyond,anderson2020rxr}. Trained methods have progressed from
waypoint-based models with panoramic RGB-D
inputs~\cite{hong2022bridging,an2025etpnav} to video-based models with a monocular
camera~\cite{zhang2024navid,zhang2025uninavid,cheng2025navila,wei2026streamvln} and navigation foundation models trained
on millions of samples~\cite{zhang2025embodied,omninav,qwennav}. Training-free methods instead build workflows
around an LLM, turning observations into text descriptions, maps or candidate waypoints to choose
from~\cite{zhou2024navgpt,chen2024mapgpt,long2024instructnav,qiao2025opennav,shi2025smartway,dai2026history}.
Both lines fit the model to the task, whereas NavRSI trains no model and lets a frontier LLM navigate with
one RGB camera and primitive actions.

\noindent\textbf{Harness-based navigation.}
As frontier LLMs learn to act in closed loop through tool calls, the design effort moves from the workflow to the
harness~\cite{zhong2026ai}. MIP~\cite{zhou2026embodied} gives the LLM a minimal interface for observing and stepping, and
runs it in harnesses ported from coding agents, such as Codex CLI and Mini-swe-agent. Later methods
design the harness by hand for navigation. They expose auxiliary tools such as depth queries, target grounding and
waypoint execution~\cite{li2026agenticnav,chen2026harnessvln,zhang2026supernav,sheng2026ask}, maintain structured memory
within or across episodes~\cite{li2026agenticnav,chen2026harnessvln,zhao2026navharness,sheng2026ask}, or
split navigation among several agent roles~\cite{shi2026navharness}, while NavGPT-3~\cite{zhou2026navgpt3} and ASENA~\cite{cheng2026asena} place the LLM
above a trained navigation model. In all of them, the harness structure is decided in advance from human priors. In contrast, NavRSI adds no tool, structured memory or agent role to the harness, and improves it from the LLM's own closed-loop behavior.

\noindent\textbf{Recursive self-improvement.}
For language agents, an LLM rewrites the
scaffold that calls it~\cite{zelikman2023self}, its own agent code~\cite{zhang2025darwin} or the harness of a
model~\cite{lee2026meta,xia2026rrsi}, and related methods optimize prompts, workflows~\cite{hu2024automated} or the search itself~\cite{zheng2026dream}. In embodied AI,
AAS~\cite{zhou2026automating} searches agent architectures, and AdaRSI~\cite{wang2026adarsi} improves a manipulation system
at the harness and policy levels. In navigation, RSI has so far been data-centric or model-centric: RSI-Nav~\cite{li2026rsinav} lets a model
generate its own training data and retrains it in each round, as do~\cite{xie2026riwanav,hong2026simplevln}, and NavGPT-3~\cite{zhou2026navgpt3} evolves the tools around a trained
model. NavRSI is harness-centric instead: it trains nothing, keeps the LLM frozen and brings such code-level
RSI~\cite{zhang2025darwin,lee2026meta} to embodied navigation under their evolve and held-out protocol~\cite{lee2026meta,xia2026rrsi}, with two designs. First, while these methods search from
the raw record of earlier candidates, our designer also distills its experience of rewriting, a compact memory that
co-evolves with the harness and can stand in for the full record. Second, the designer reads simulator diagnostics
that the navigator never sees, so the harness needs no privileged information.

\section{Method}
\label{sec:method}

\begin{figure*}[t]
\centering
\includegraphics[width=\textwidth]{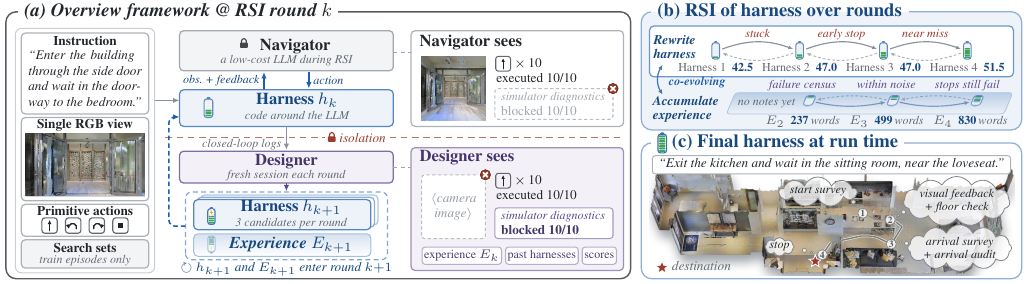}
\caption{%
Overview of NavRSI. (a)~One RSI round: in harness $h_k$, the frozen navigator runs training episodes, and the designer reads the closed-loop
logs and its experience $E_k$ to write $h_{k+1}$ and $E_{k+1}$. An isolation wall splits what
the two see at the same step: the navigator gets the camera view but no simulator diagnostics, and the designer gets the
diagnostics but no image. (b)~Search-set SR of each round's best harness, and the growing experience. (c)~The final harness for evaluation.}
\label{fig:overview}
\end{figure*}

As shown in \cref{fig:overview}, NavRSI keeps the navigator frozen and recursively rewrites only the harness around
it, driven by the navigator's own closed-loop behavior. After the setup (\S\ref{sec:setup}), we present
the two parts of co-evolving RSI, \textit{harness rewriting} (\S\ref{sec:rsi}) and \textit{experience accumulation}
(\S\ref{sec:dual}), then \textit{information
isolation} (\S\ref{sec:isolation}) and the
\textit{final harness} (\S\ref{sec:harness}).

\subsection{Problem setup}
\label{sec:setup}

In VLN-CE~\cite{krantz2020beyond}, a robot is placed in an unseen environment and receives a language instruction $I$. In our setting, it perceives the environment only through a single forward-facing RGB camera and acts with primitive actions, \textit{e.g.}, move
forward 0.25\,m, turn left or right by 15$^\circ$, or STOP. A frozen LLM $\mathcal{M}$, the \emph{navigator}, drives the robot through a \emph{harness} $h$,
\ie, the program between the navigator and the environment that decides its context, tools and
feedback~\cite{zhong2026ai}. For an episode with instruction $I$ in environment $e$, the harness runs the closed loop and
returns the trajectory
\begin{equation}
\tau = h\left(I, e;\, \mathcal{M}\right),
\label{eq:harness}
\end{equation}
where the navigator acts through two tools only: \textit{observe} returns the current camera view, and \textit{step} executes a
list of primitive actions. With $\mathcal{M}$ fixed, we seek the harness that succeeds most often in unseen environments,
\begin{equation}
h^{*} = \arg\max_{h}\; \mathbb{E}_{(I,e)\sim\mathcal{D}_{\mathrm{unseen}}}\left[ s\left(h(I,e;\,\mathcal{M})\right) \right],
\label{eq:objective}
\end{equation}
where $s(\tau)\in\{0,1\}$ marks success. \textbf{\textit{Note that the RSI search uses only training episodes.}}

\subsection{Harness rewriting via RSI}
\label{sec:rsi}

RSI starts from the open-sourced Mini-swe-agent harness $h_0$~\cite{zhou2026embodied} (Seed~0 in \cref{tab:r2r_mip}) and rewrites it round by
round (\cref{fig:overview}(a)). Each round $k$ consists of three steps, \ie, \textbf{\textit{closed-loop rollout}},
\textbf{\textit{designer rewriting}} and \textbf{\textit{staged evaluation}}.

\paragraph{Closed-loop rollout.} During RSI, a low-cost LLM $\mathcal{M}_{\ell}$ stands in for $\mathcal{M}$. It runs the
current harness $h_k$ on a search set $\mathcal{D}_{S}$ of training episodes, yielding the closed-loop logs
\begin{equation}
\mathcal{L}_k = \left\{\, h_k\left(I, e;\, \mathcal{M}_{\ell}\right) \;\middle|\; (I,e)\in\mathcal{D}_{S} \,\right\},
\label{eq:logs}
\end{equation}
which record every model call, tool call and harness message. Meanwhile, the simulator records diagnostics $\Omega_k$ that the robot never has: after every \textit{step} call, the distance to the goal, the progress along the reference path and the blocked moves, and for each episode a failure type such as stuck, early stop or near miss.

\paragraph{Designer rewriting.} A designer agent $\mathcal{A}$, a frontier LLM working in a fresh session every round,
reads the logs, the diagnostics, all harnesses evaluated so far $\mathcal{H}_k$ with their search-set success rates
$\mathrm{SR}_S(\cdot)$, and its experience $E_k$ (\S\ref{sec:dual}). It then writes $C$ candidate harnesses as code and updates its experience for the next round,
\begin{equation}
\left(\{h_{k+1}^{(c)}\}_{c=1}^{C},\; E_{k+1}\right) = \mathcal{A}\left(\mathcal{L}_k,\, \Omega_k,\, \mathcal{H}_k,\, E_k\right).
\label{eq:designer}
\end{equation}
The designer may rewrite any part of the harness, \eg, the prompts, the agent loop, and what the navigator sees and when, but not the navigator, the tools or the budgets.

\paragraph{Staged evaluation.} Each candidate must pass a rule check and a dry run, and then runs on the search set
in stages, stopping early once it falls clearly behind the best harness so far. The best harness after round $k$ is
$h_{k+1} = \arg\max_{h} \mathrm{SR}_{S}(h)$ over all harnesses evaluated so far,
and the designer builds the next round on the strongest ones. Finally, we select the final harness on development episodes. Such closed-loop rewriting is essential: harnesses written in one shot by the same designer all fall below Seed~0 (\cref{tab:abl_oneshot}).

\subsection{Experience accumulation}
\label{sec:dual}

As each round starts a fresh designer session, the designer's reasoning would be lost between rounds: past harnesses and scores show what was tried, but not why or what it taught. NavRSI thus keeps the designer's
\textbf{\textit{experience of rewriting}} $E_k$, which co-evolves with the harness.

\paragraph{Written experience.} $E_k$ holds working notes that the designer writes and revises every round, in four parts: a \textit{failure census}, the \textit{hypothesis} behind each rewrite,
\textit{effect checks} against run-to-run noise, and \textit{next steps} to try or to avoid.

\paragraph{Co-evolution.} Each rewrite is judged in closed loop, its outcome enters the experience, and the experience
steers the next rewrite, so the harness and the experience co-evolve (\cref{fig:overview}(b), \cref{fig:evolution}). This experience keeps RSI from repeating
itself: with no record of earlier rounds, 52.8\% of the designer's new rewrites repeat an earlier one, but none do with
$E_k$ alone (\cref{tab:abl_exp}). \textbf{\textit{RSI thus improves not only the harness, but
also the designer's way of rewriting it.}}

\subsection{Information isolation}
\label{sec:isolation}

NavRSI separates what each agent may see (\cref{fig:overview}(a), right): \textbf{\textit{the navigator never sees the
simulator diagnostics, and the designer never sees an image}}.
\begin{itemize}
\item \textit{Navigator side:} in every rollout and at test time, the navigator receives only the instruction, its camera
views and the harness's feedback, all of which a real robot has.
\item \textit{Designer side:} the designer reads training logs in which every image is replaced by a placeholder, with the diagnostics, the past harnesses, their scores and its experience. It never sees a development or
test episode, and each session starts from one fixed briefing on the interface, the budgets, the rules and general design hints.
\item \textit{Rules and data:} episodes run in fresh processes, a rule check flags any code that embeds episode content, and the search sets come from R2R-CE training scenes only.
\end{itemize}
Unlike prior work that evolves its agent on evaluation sets~\cite{zhou2026automating,cheng2026asena}, NavRSI thus searches on training episodes only.

\setcounter{topnumber}{1}
\providecommand{\flame}{\,\raisebox{-0.14em}{\includegraphics[height=0.95em]{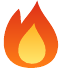}}}
\begin{table*}[t]
  \centering
  \footnotesize
  \caption{\textbf{Comparison on the full R2R-CE and RxR-CE Val-Unseen splits}.
  Pano.: panoramic or multi-view RGB. Odom.: odometry or pose. S.RGB: single RGB camera. External Data: training data from outside
  the simulators. ``--'': not reported. \flame: trained model. \textbf{\textit{Please note that NavRSI's RxR-CE results are direct
  cross-dataset generalization from R2R-CE's harness, with no tuning on RxR-CE.}}}
  \label{tab:r2r_full}
  \label{tab:rxr500}
  \setlength{\tabcolsep}{2.7pt} %
  \setlength{\aboverulesep}{0pt}
  \setlength{\belowrulesep}{0pt}
  \renewcommand{\arraystretch}{1.1} %
  \setlength{\dashlinedash}{2pt}
  \setlength{\dashlinegap}{1.5pt}
  \begin{tabular}{lcccccccccccccc}
    \specialrule{1pt}{0pt}{0pt}
    \rowcolor[gray]{0.9} & \multicolumn{4}{c}{Observation} & Training & \multicolumn{4}{c}{R2R-CE Val-Unseen} & \multicolumn{4}{c}{RxR-CE Val-Unseen} \\
    \noalign{\vskip-\cmidrulewidth}\cmidrule(lr){2-5}\cmidrule(lr){6-6}\cmidrule(lr){7-10}\cmidrule(lr){11-14}
    \rowcolor[gray]{0.9}\multirow{-2}{*}{Method} & Pano. & Odom. & Depth & S.RGB & External Data & NE$\downarrow$ & OSR$\uparrow$ & SR$\uparrow$ & SPL$\uparrow$ & NE$\downarrow$ & nDTW$\uparrow$ & SR$\uparrow$ & SPL$\uparrow$ \\
    \specialrule{0.5pt}{0pt}{1.2pt}\specialrule{0.5pt}{0pt}{0pt}
    Uni-NaVid~\cite{zhang2025uninavid}\flame &  &  &  & \checkmark & 2.30M & 5.58 & 53.3 & 47.0 & 42.7 & 6.24 & -- & 48.7 & 40.9 \\
    NaVILA~\cite{cheng2025navila}\flame &  &  &  & \checkmark & 2.22M & 5.22 & 62.5 & 54.0 & 49.0 & 6.77 & -- & 49.3 & 44.0 \\
    Aux-Think~\cite{wang2025auxthink}\flame &  &  &  & \checkmark & 1.60M & 6.08 & 60.0 & 54.8 & 46.9 & 6.24 & -- & 52.2 & 40.2 \\
    RSI-Nav~\cite{li2026rsinav}\flame &  &  &  & \checkmark & -- & \mc{11}{3.70} & \mc{13}{78.0} & \mc{10}{68.6} & \mc{6}{58.6} & \mc{12}{3.53} & \mc{12}{70.6} & \mc{10}{72.4} & 56.5 \\
    NavFoM~\cite{zhang2025embodied}\flame & \checkmark &  &  &  & 7.47M & 4.61 & \mc{2}{72.1} & 61.7 & 55.3 & 4.74 & 65.8 & 64.4 & 56.2 \\
    ABot-N0~\cite{abotN0}\flame & \checkmark &  &  &  & 8.44M & \mc{10}{3.78} & 70.8 & \mc{7}{66.4} & \mc{14}{63.9} & \mc{9}{3.83} & -- & \mc{6}{69.3} & \mc{6}{60.0} \\
    AstraNav-World~\cite{astranavworld}\flame & \checkmark &  &  &  & -- & \mc{9}{3.86} & \mc{6}{73.9} & \mc{9}{67.9} & \mc{16}{65.4} & \mc{9}{3.82} & -- & \mc{10}{72.9} & \mc{9}{61.5} \\
    OmniNav~\cite{omninav}\flame & \checkmark &  &  &  & 9.20M & \mc{11}{3.74} & \mc{7}{74.6} & \mc{12}{69.5} & \mc{17}{66.1} & \mc{9}{3.77} & -- & \mc{11}{73.6} & \mc{10}{62.0} \\
    Qwen-RobotNav~\cite{qwennav}\flame & \checkmark &  &  &  & 4.26M & \mc{13}{3.53} & \mc{14}{78.5} & \mc{16}{72.1} & \mc{18}{66.6} & \mc{11}{3.58} & \mc{16}{72.5} & \mc{14}{76.5} & \mc{16}{65.7} \\
    NavGPT VLA~\cite{zhou2026navgpt3}\flame & \checkmark &  &  &  & 6.71M & \mc{20}{2.97} & \mc{22}{82.49} & \mc{23}{76.94} & \mc{24}{70.61} & \mc{20}{2.65} & \mc{26}{76.74} & \mc{21}{82.39} & \mc{24}{70.56} \\
    \textsc{NavGPT-3}~\cite{zhou2026navgpt3} (GPT-6-Astra + 8B VLA\flame) & \checkmark & \checkmark & \checkmark &  & 6.71M & \mc{30}{2.18} & \mc{30}{86.99} & \mc{30}{81.51} & \mc{24}{70.42} & \mc{30}{\textbf{1.56}} & \mc{30}{78.47} & \mc{30}{\textbf{90.43}} & \mc{30}{\textbf{74.31}} \\
    \hdashline
    \textbf{NavRSI (ours)} (GPT-6-Astra) &  &  &  & \checkmark & \textbf{0} & \mc{31}{\textbf{2.11}} & \mc{35}{\textbf{89.45}} & \mc{36}{\textbf{85.64}} & \mc{29}{\textbf{73.58}} & \mc{28}{1.78} & \mc{31}{\textbf{78.97}} & \mc{28}{88.96} & \mc{26}{71.99} \\
    \specialrule{1pt}{0pt}{0pt}
  \end{tabular}
\end{table*}

\begin{table}[t]
\caption{\textbf{Comparison on NavRAG-CE}~\cite{wang2025navrag}. Above the line: trained models from
D3D-VLP~\cite{wang2026d3dvlp} (full set). Below: 100 val-unseen episodes, GPT-6-Astra, mean $\pm$ s.d. of three runs.
\textbf{\textit{Please note that NavRSI is evaluated in cross-dataset generalization.}}}
\label{tab:navrag}
\centering
\scriptsize
\setlength{\tabcolsep}{2pt}%
\setlength{\aboverulesep}{0pt}
\setlength{\belowrulesep}{0pt}
\renewcommand{\arraystretch}{1.45}%
\setlength{\dashlinedash}{2pt}
\setlength{\dashlinegap}{1.5pt}
\resizebox{\columnwidth}{!}{%
\begin{tabular}{lcccc}
\specialrule{1pt}{0pt}{0pt}
\rowcolor[gray]{0.9}Method & NE$\downarrow$ & OSR$\uparrow$ & SR$\uparrow$ & SPL$\uparrow$ \\
\specialrule{0.5pt}{0pt}{1.2pt}\specialrule{0.5pt}{0pt}{0pt}
NaVid~\cite{zhang2024navid} & \cellcolor[rgb]{0.984,0.992,1.000}9.35 & \cellcolor[rgb]{0.984,0.992,1.000}29.6 & \cellcolor[rgb]{0.984,0.992,1.000}19.4 & \cellcolor[rgb]{0.984,0.992,1.000}13.9 \\
Dynam3D~\cite{wang2025dynam3d} & \cellcolor[rgb]{0.984,0.992,1.000}8.12 & \cellcolor[rgb]{0.984,0.992,1.000}38.4 & \cellcolor[rgb]{0.984,0.992,1.000}24.7 & \cellcolor[rgb]{0.984,0.992,1.000}18.8 \\
D3D-VLP~\cite{wang2026d3dvlp} & \cellcolor[rgb]{0.984,0.992,1.000}7.57 & \cellcolor[rgb]{0.984,0.992,1.000}45.3 & \cellcolor[rgb]{0.984,0.992,1.000}31.1 & \cellcolor[rgb]{0.979,0.989,1.000}23.9 \\
\midrule
CA-Nav~\cite{chen2025canav} (GPT-6-Astra) & \cellcolor[rgb]{0.984,0.992,1.000}8.57{\scriptsize$\,\pm$0.27} & \cellcolor[rgb]{0.984,0.992,1.000}51.7{\scriptsize$\,\pm$1.2} & \cellcolor[rgb]{0.984,0.992,1.000}11.3{\scriptsize$\,\pm$3.5} & \cellcolor[rgb]{0.984,0.992,1.000}2.8{\scriptsize$\,\pm$1.2} \\
Open-Nav~\cite{qiao2025opennav} (GPT-6-Astra) & \cellcolor[rgb]{0.978,0.989,1.000}7.00{\scriptsize$\,\pm$0.20} & \cellcolor[rgb]{0.984,0.992,1.000}48.3{\scriptsize$\,\pm$1.5} & \cellcolor[rgb]{0.984,0.992,1.000}23.7{\scriptsize$\,\pm$3.2} & \cellcolor[rgb]{0.984,0.992,1.000}14.5{\scriptsize$\,\pm$1.5} \\
SmartWay~\cite{shi2025smartway} (GPT-6-Astra) & \cellcolor[rgb]{0.778,0.886,0.995}\textbf{6.24}{\scriptsize$\,\pm$0.26} & \cellcolor[rgb]{0.984,0.992,1.000}56.3{\scriptsize$\,\pm$2.1} & \cellcolor[rgb]{0.984,0.992,1.000}33.3{\scriptsize$\,\pm$2.5} & \cellcolor[rgb]{0.984,0.992,1.000}21.1{\scriptsize$\,\pm$1.5} \\
O2C-Nav~\cite{pan2026one} (GPT-6-Astra) & \cellcolor[rgb]{0.984,0.992,1.000}7.14{\scriptsize$\,\pm$0.18} & \cellcolor[rgb]{0.951,0.975,0.999}60.7{\scriptsize$\,\pm$3.1} & \cellcolor[rgb]{0.984,0.992,1.000}31.3{\scriptsize$\,\pm$2.5} & \cellcolor[rgb]{0.984,0.992,1.000}16.8{\scriptsize$\,\pm$0.5} \\
MIP~\cite{zhou2026embodied} (GPT-6-Astra) & \cellcolor[rgb]{0.984,0.992,1.000}7.41{\scriptsize$\,\pm$0.14} & \cellcolor[rgb]{0.873,0.935,0.997}64.0{\scriptsize$\,\pm$4.4} & \cellcolor[rgb]{0.982,0.991,1.000}34.3{\scriptsize$\,\pm$3.8} & \cellcolor[rgb]{0.984,0.992,1.000}21.3{\scriptsize$\,\pm$1.1} \\
\textsc{NavGPT-3}~\cite{zhou2026navgpt3} (GPT-6-Astra + 8B VLA\flame) & \cellcolor[rgb]{0.984,0.992,1.000}8.00{\scriptsize$\,\pm$0.22} & \cellcolor[rgb]{0.886,0.942,0.997}63.5{\scriptsize$\,\pm$1.4} & \cellcolor[rgb]{0.984,0.992,1.000}29.4{\scriptsize$\,\pm$2.7} & \cellcolor[rgb]{0.984,0.992,1.000}14.8{\scriptsize$\,\pm$0.8} \\
\hdashline
\textbf{NavRSI (ours)} (GPT-6-Astra) & \cellcolor[rgb]{0.977,0.988,0.999}7.00{\scriptsize$\,\pm$0.26} & \cellcolor[rgb]{0.778,0.886,0.995}\textbf{66.7}{\scriptsize$\,\pm$1.5} & \cellcolor[rgb]{0.778,0.886,0.995}\textbf{39.7}{\scriptsize$\,\pm$5.0} & \cellcolor[rgb]{0.778,0.886,0.995}\textbf{27.4}{\scriptsize$\,\pm$1.3} \\
\specialrule{1pt}{0pt}{0pt}
\end{tabular}%
}
\end{table}

\subsection{The final harness}
\label{sec:harness}

\cref{fig:overview}(c) shows the final harness on a test episode. RSI turns Seed~0 into a harness with three abilities, whose six modules M1--M6 (\cref{tab:abl_arms}) were all written by the designer.
\begin{itemize}
\item \textit{Action feedback:} \emph{visual feedback} (M1) compares consecutive camera views in code and, after a forward batch with a
nearly unchanged view, reports a likely obstruction (``executed actions are not measured travel''). A \emph{floor check} (M4) then tilts the camera down to the floor.
\item \textit{Progress tracking:} before the first move, a \emph{start survey} (M2) turns the robot around once and gives the
navigator four labeled views to relate the first route segment to its surroundings. M1 also labels every later view with the
heading and camera pitch implied by the executed actions, and flags a view that nearly matches an earlier one at the same
heading as a revisit.
\item \textit{Arrival verification:} the first STOP becomes a proposal. An \emph{arrival survey} (M3) takes fresh views
around the robot, and an \emph{arrival audit} (M5), run one reasoning-effort level higher (M6), asks the
navigator to compare the destination with similar places along the route and to stop, move closer or go back, with at
most two corrections.
\end{itemize}
\textbf{\textit{The final harness adds no tool, structured memory or agent role, and every movement decision stays with the navigator}}: the harness only turns or tilts the camera in place to gather views, and keeps past views only to compare them within an episode. Found with the low-cost $\mathcal{M}_{\ell}$, it runs unchanged with LLMs from four vendors
(\cref{tab:r2r_mip,tab:r2r_native}).

\section{Experiments}
\label{sec:experiments}

\providecommand{\rqtag}[1]{\tikz[baseline=(rq.base)]\node[draw=black!75,fill=black!4,rounded corners=1.5pt,%
  line width=0.45pt,inner xsep=2.1pt,inner ysep=1.2pt,outer sep=0pt](rq){\sffamily\bfseries\scriptsize RQ#1};}
\makeatletter
\providecommand{\rqfont}{\normalfont\bfseries\hyphenpenalty=\@M\exhyphenpenalty=\@M}
\newlength{\rq@wd}
\newcommand{\rq@section}{\@startsection{paragraph}{4}{\z@}{-5pt plus -2pt minus -1pt}{2pt plus 1pt}{\rqfont}}
\newcommand{\rq@hang}[1]{\settowidth{\rq@wd}{\rqtag{#1}\hspace{0.45em}}\hangindent=\rq@wd\hangafter=1\relax}
\newcommand{\rq@needspace}[1]{\par\penalty-100\begingroup\dimen@=#1\relax\dimen@ii=\pagegoal\advance\dimen@ii-\pagetotal
  \ifdim\dimen@>\dimen@ii\ifdim\dimen@ii>\z@\vfil\fi\break\fi\endgroup}
\providecommand{\rqsec}[2]{\rq@needspace{4\baselineskip}\rq@section*{\rq@hang{#1}\rqtag{#1}\hspace{0.45em}#2}}
\makeatother

\providecommand{\flame}{\,\raisebox{-0.14em}{\includegraphics[height=0.95em]{figures/icons/flame.pdf}}}
\begin{table}[t]
\caption{\textbf{Efficiency on RxR-CE.} Per-episode means on the same 800 val-unseen episodes (GPT-6-Astra). Time and GPU Mem. are measured on the same machine (RTX 5000) for a fair comparison.}
\label{tab:rxr500_eff}
\centering
\footnotesize
\setlength{\tabcolsep}{1.6pt}
\setlength{\aboverulesep}{0pt}
\setlength{\belowrulesep}{0pt}
\renewcommand{\arraystretch}{1.2}
\setlength{\dashlinedash}{2pt}
\setlength{\dashlinegap}{1.5pt}
\resizebox{\columnwidth}{!}{%
\begin{tabular}{lcccccc}
\specialrule{1pt}{0pt}{0pt}
\rowcolor[gray]{0.9} & \multicolumn{2}{c}{Navigation} & \multicolumn{2}{c}{API use} & \multicolumn{2}{c}{GPU use} \\
\noalign{\vskip-\cmidrulewidth}\cmidrule(lr){2-3}\cmidrule(lr){4-5}\cmidrule(lr){6-7}
\rowcolor[gray]{0.9}\multirow{-2}{*}{Method} & SR$\uparrow$ & SPL$\uparrow$ & Tok.\ (k)$\downarrow$ & Cost (\$)$\downarrow$ & Time (s)$\downarrow$ & Mem.\ (GB)$\downarrow$ \\
\specialrule{0.5pt}{0pt}{1.2pt}\specialrule{0.5pt}{0pt}{0pt}
\textsc{NavGPT-3}~\cite{zhou2026navgpt3} + 8B VLA\flame & \cellcolor[rgb]{0.793,0.894,0.995}88.75 & \cellcolor[rgb]{0.804,0.900,0.996}72.57 & \cellcolor[rgb]{0.988,0.769,0.832}\textbf{235.8} & \cellcolor[rgb]{0.997,0.939,0.955}0.562 & \cellcolor[rgb]{0.991,0.814,0.864}333.2 & \cellcolor[rgb]{0.999,0.984,0.988}18.97 \\
MIP~\cite{zhou2026embodied} (Codex CLI) & \cellcolor[rgb]{0.835,0.916,0.996}87.25 & \cellcolor[rgb]{0.810,0.903,0.996}72.39 & \cellcolor[rgb]{0.999,0.984,0.988}706.8 & \cellcolor[rgb]{0.999,0.984,0.988}1.086 & \cellcolor[rgb]{0.998,0.957,0.969}434.8 & \cellcolor[rgb]{0.988,0.769,0.832}\textbf{0.93} \\
\hdashline
\textbf{NavRSI (ours)} & \cellcolor[rgb]{0.778,0.886,0.995}\textbf{89.25} & \cellcolor[rgb]{0.778,0.886,0.995}\textbf{73.30} & \cellcolor[rgb]{0.991,0.824,0.872}255.1 & \cellcolor[rgb]{0.988,0.769,0.832}\textbf{0.424} & \cellcolor[rgb]{0.988,0.769,0.832}\textbf{312.8} & \cellcolor[rgb]{0.988,0.769,0.832}\textbf{0.93} \\
\specialrule{1pt}{0pt}{0pt}
\end{tabular}%
}
\end{table}

\begin{table*}[t]
\caption{\textbf{Comparison with each LLM's native harness on R2R-CE-100}~\cite{qiao2025opennav}, at each LLM's default API reasoning effort (NavRSI's audit one level up, M6). All NavRSI rows run the same final
harness, and only the LLM changes. Native rows are from
MIP~\cite{zhou2026embodied} or our runs. Navigation is mean $\pm$ s.d. of three runs. Cost/SR and Tokens/SR are per successful episode at list prices. Darker shading is better.}
\label{tab:r2r_native}
\centering
\footnotesize
\providecommand{\harnessicon}[1]{\makebox[1.23em][l]{#1}}
\providecommand{\iconclaude}{\raisebox{-0.16em}{\includegraphics[height=0.95em]{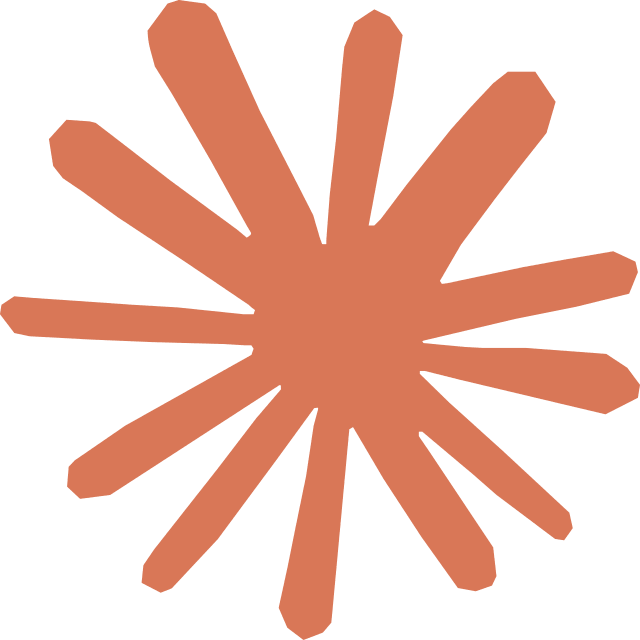}}}
\providecommand{\icongemini}{\raisebox{-0.16em}{\includegraphics[height=0.95em]{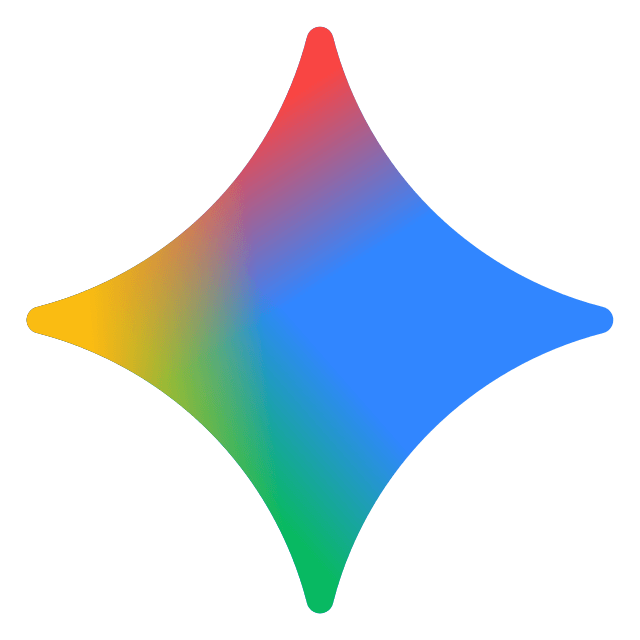}}}
\providecommand{\iconopenai}{\raisebox{-0.16em}{\includegraphics[height=0.95em]{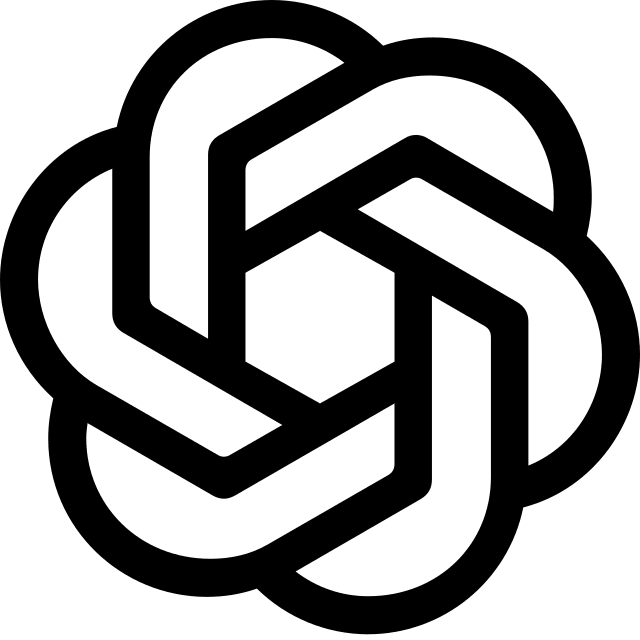}}}
\providecommand{\iconqwen}{\raisebox{-0.16em}{\includegraphics[height=0.95em]{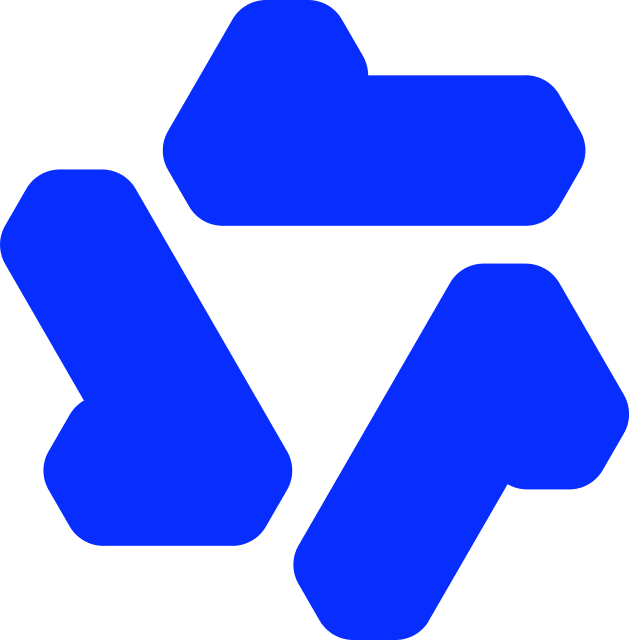}}}
\setlength{\tabcolsep}{3.6pt}
\setlength{\aboverulesep}{0pt}
\setlength{\belowrulesep}{0pt}
\renewcommand{\arraystretch}{1.2}
\setlength{\dashlinedash}{2pt}
\setlength{\dashlinegap}{1.5pt}
\resizebox{\textwidth}{!}{%
\begin{tabular}{llccccccccc}
\specialrule{1pt}{0pt}{0pt}
\rowcolor[gray]{0.9} & & \multicolumn{4}{c}{Navigation} & \multicolumn{5}{c}{Resource use} \\
\noalign{\vskip-\cmidrulewidth}\cmidrule(lr){3-6}\cmidrule(lr){7-11}
\rowcolor[gray]{0.9}\multirow{-2}{*}{Harness} & \multirow{-2}{*}{LLMs} & NE$\downarrow$ & OSR$\uparrow$ & SR$\uparrow$ & SPL$\uparrow$ & Turns$\downarrow$ & Cost (\$)$\downarrow$ & Tokens (k)$\downarrow$ & Tokens/SR (k)$\downarrow$ & Cost/SR (\$)$\downarrow$ \\
\specialrule{0.5pt}{0pt}{1.2pt}\specialrule{0.5pt}{0pt}{0pt}
\harnessicon{\iconclaude}Claude Agent SDK & sonnet-5 & \cellcolor[rgb]{0.984,0.992,1.000}5.80{\scriptsize$\,\pm$0.43} & \cellcolor[rgb]{0.876,0.937,0.997}61.3{\scriptsize$\,\pm$1.2} & \cellcolor[rgb]{0.972,0.986,0.999}51.3{\scriptsize$\,\pm$1.2} & \cellcolor[rgb]{0.984,0.992,1.000}37.8{\scriptsize$\,\pm$0.9} & \cellcolor[rgb]{0.992,0.832,0.878}95 & \cellcolor[rgb]{0.999,0.984,0.988}3.01 & \cellcolor[rgb]{0.994,0.883,0.915}2816 & \cellcolor[rgb]{0.997,0.937,0.954}5489 & \cellcolor[rgb]{0.999,0.984,0.988}5.867 \\
\hdashline
\harnessicon{}\textbf{NavRSI (ours)} & sonnet-5 & \cellcolor[rgb]{0.809,0.903,0.996}\textbf{5.08}{\scriptsize$\,\pm$0.49} & \cellcolor[rgb]{0.809,0.903,0.996}\textbf{63.3}{\scriptsize$\,\pm$3.2} & \cellcolor[rgb]{0.809,0.903,0.996}\textbf{57.7}{\scriptsize$\,\pm$1.5} & \cellcolor[rgb]{0.809,0.903,0.996}\textbf{45.7}{\scriptsize$\,\pm$1.2} & \cellcolor[rgb]{0.990,0.802,0.856}\textbf{90.3} & \cellcolor[rgb]{0.990,0.802,0.856}\textbf{0.858} & \cellcolor[rgb]{0.990,0.802,0.856}\textbf{2443} & \cellcolor[rgb]{0.990,0.802,0.856}\textbf{4236} & \cellcolor[rgb]{0.990,0.802,0.856}\textbf{1.487} \\
\midrule
\harnessicon{\iconclaude}Claude Agent SDK & opus-5 & \cellcolor[rgb]{0.984,0.992,1.000}4.79{\scriptsize$\,\pm$0.43} & \cellcolor[rgb]{0.855,0.926,0.997}78.3{\scriptsize$\,\pm$5.7} & \cellcolor[rgb]{0.934,0.966,0.999}70.7{\scriptsize$\,\pm$3.5} & \cellcolor[rgb]{0.969,0.984,0.999}55.2{\scriptsize$\,\pm$2.7} & \cellcolor[rgb]{0.995,0.898,0.926}53 & \cellcolor[rgb]{0.996,0.910,0.935}1.27 & \cellcolor[rgb]{0.999,0.984,0.988}1283 & \cellcolor[rgb]{0.999,0.984,0.988}1815 & \cellcolor[rgb]{0.997,0.941,0.957}1.796 \\
\hdashline
\harnessicon{}\textbf{NavRSI (ours)} & opus-5 & \cellcolor[rgb]{0.809,0.903,0.996}\textbf{3.66}{\scriptsize$\,\pm$0.39} & \cellcolor[rgb]{0.809,0.903,0.996}\textbf{80.0}{\scriptsize$\,\pm$1.7} & \cellcolor[rgb]{0.809,0.903,0.996}\textbf{76.0}{\scriptsize$\,\pm$1.0} & \cellcolor[rgb]{0.809,0.903,0.996}\textbf{61.7}{\scriptsize$\,\pm$2.2} & \cellcolor[rgb]{0.990,0.802,0.856}\textbf{44.6} & \cellcolor[rgb]{0.990,0.802,0.856}\textbf{1.042} & \cellcolor[rgb]{0.990,0.802,0.856}\textbf{795} & \cellcolor[rgb]{0.990,0.802,0.856}\textbf{1046} & \cellcolor[rgb]{0.990,0.802,0.856}\textbf{1.371} \\
\midrule
\harnessicon{\iconclaude}Claude Agent SDK & opus-5.5 & \cellcolor[rgb]{0.981,0.990,1.000}3.03{\scriptsize$\,\pm$0.51} & \cellcolor[rgb]{0.826,0.911,0.996}84.7{\scriptsize$\,\pm$3.2} & \cellcolor[rgb]{0.829,0.912,0.996}82.3{\scriptsize$\,\pm$3.5} & \cellcolor[rgb]{0.831,0.914,0.996}71.3{\scriptsize$\,\pm$2.7} & \cellcolor[rgb]{0.994,0.890,0.920}34 & \cellcolor[rgb]{0.991,0.820,0.869}0.21 & \cellcolor[rgb]{0.998,0.951,0.964}232 & \cellcolor[rgb]{0.998,0.954,0.966}282 & \cellcolor[rgb]{0.991,0.825,0.872}0.255 \\
\hdashline
\harnessicon{}\textbf{NavRSI (ours)} & opus-5.5 & \cellcolor[rgb]{0.809,0.903,0.996}\textbf{2.68}{\scriptsize$\,\pm$0.38} & \cellcolor[rgb]{0.809,0.903,0.996}\textbf{85.3}{\scriptsize$\,\pm$3.1} & \cellcolor[rgb]{0.809,0.903,0.996}\textbf{83.0}{\scriptsize$\,\pm$1.7} & \cellcolor[rgb]{0.809,0.903,0.996}\textbf{72.0}{\scriptsize$\,\pm$2.5} & \cellcolor[rgb]{0.990,0.802,0.856}\textbf{29.1} & \cellcolor[rgb]{0.990,0.802,0.856}\textbf{0.204} & \cellcolor[rgb]{0.990,0.802,0.856}\textbf{172} & \cellcolor[rgb]{0.990,0.802,0.856}\textbf{208} & \cellcolor[rgb]{0.990,0.802,0.856}\textbf{0.246} \\
\midrule
\harnessicon{\iconclaude}Claude Agent SDK & fable-5 & \cellcolor[rgb]{0.984,0.992,1.000}5.13{\scriptsize$\,\pm$0.26} & \cellcolor[rgb]{0.897,0.947,0.998}73.3{\scriptsize$\,\pm$1.5} & \cellcolor[rgb]{0.921,0.960,0.998}68.3{\scriptsize$\,\pm$1.5} & \cellcolor[rgb]{0.939,0.969,0.999}58.0{\scriptsize$\,\pm$1.5} & \cellcolor[rgb]{0.999,0.977,0.984}63 & \cellcolor[rgb]{0.999,0.984,0.988}3.43 & \cellcolor[rgb]{0.999,0.984,0.988}1732 & \cellcolor[rgb]{0.999,0.984,0.988}2536 & \cellcolor[rgb]{0.999,0.984,0.988}5.022 \\
\hdashline
\harnessicon{}\textbf{NavRSI (ours)} & fable-5 & \cellcolor[rgb]{0.809,0.903,0.996}\textbf{4.05}{\scriptsize$\,\pm$0.17} & \cellcolor[rgb]{0.809,0.903,0.996}\textbf{76.7}{\scriptsize$\,\pm$3.1} & \cellcolor[rgb]{0.809,0.903,0.996}\textbf{72.7}{\scriptsize$\,\pm$3.8} & \cellcolor[rgb]{0.809,0.903,0.996}\textbf{62.6}{\scriptsize$\,\pm$3.6} & \cellcolor[rgb]{0.990,0.802,0.856}\textbf{42.3} & \cellcolor[rgb]{0.990,0.802,0.856}\textbf{1.856} & \cellcolor[rgb]{0.990,0.802,0.856}\textbf{744} & \cellcolor[rgb]{0.990,0.802,0.856}\textbf{1024} & \cellcolor[rgb]{0.990,0.802,0.856}\textbf{2.554} \\
\midrule
\harnessicon{\iconclaude}Claude Agent SDK & fable-5.2 & \cellcolor[rgb]{0.970,0.985,0.999}2.37{\scriptsize$\,\pm$0.15} & \cellcolor[rgb]{0.897,0.947,0.998}87.0{\scriptsize$\,\pm$0.0} & \cellcolor[rgb]{0.900,0.949,0.998}83.0{\scriptsize$\,\pm$1.0} & \cellcolor[rgb]{0.893,0.945,0.998}73.7{\scriptsize$\,\pm$0.7} & \cellcolor[rgb]{0.992,0.850,0.891}21.8 & \cellcolor[rgb]{0.990,0.802,0.856}\textbf{0.193} & \cellcolor[rgb]{0.993,0.867,0.903}108 & \cellcolor[rgb]{0.995,0.892,0.921}130 & \cellcolor[rgb]{0.990,0.802,0.856}\textbf{0.233} \\
\hdashline
\harnessicon{}\textbf{NavRSI (ours)} & fable-5.2 & \cellcolor[rgb]{0.809,0.903,0.996}\textbf{2.14}{\scriptsize$\,\pm$0.22} & \cellcolor[rgb]{0.809,0.903,0.996}\textbf{91.0}{\scriptsize$\,\pm$2.0} & \cellcolor[rgb]{0.809,0.903,0.996}\textbf{87.0}{\scriptsize$\,\pm$2.6} & \cellcolor[rgb]{0.809,0.903,0.996}\textbf{76.9}{\scriptsize$\,\pm$1.6} & \cellcolor[rgb]{0.990,0.802,0.856}\textbf{20.1} & \cellcolor[rgb]{0.999,0.974,0.981}0.283 & \cellcolor[rgb]{0.990,0.802,0.856}\textbf{96} & \cellcolor[rgb]{0.990,0.802,0.856}\textbf{110} & \cellcolor[rgb]{0.998,0.962,0.973}0.325 \\
\midrule
\harnessicon{\iconopenai}Codex CLI & gpt-5.6-sol & \cellcolor[rgb]{0.984,0.992,1.000}5.34{\scriptsize$\,\pm$0.45} & \cellcolor[rgb]{0.984,0.992,1.000}61.3{\scriptsize$\,\pm$2.9} & \cellcolor[rgb]{0.984,0.992,1.000}56.7{\scriptsize$\,\pm$2.5} & \cellcolor[rgb]{0.984,0.992,1.000}41.8{\scriptsize$\,\pm$2.6} & \cellcolor[rgb]{0.992,0.831,0.877}42 & \cellcolor[rgb]{0.999,0.984,0.988}0.76 & \cellcolor[rgb]{0.999,0.984,0.988}1350 & \cellcolor[rgb]{0.999,0.984,0.988}2380 & \cellcolor[rgb]{0.999,0.984,0.988}1.340 \\
\hdashline
\harnessicon{}\textbf{NavRSI (ours)} & gpt-5.6-sol & \cellcolor[rgb]{0.809,0.903,0.996}\textbf{4.10}{\scriptsize$\,\pm$0.40} & \cellcolor[rgb]{0.809,0.903,0.996}\textbf{77.0}{\scriptsize$\,\pm$1.7} & \cellcolor[rgb]{0.809,0.903,0.996}\textbf{68.0}{\scriptsize$\,\pm$4.0} & \cellcolor[rgb]{0.809,0.903,0.996}\textbf{53.5}{\scriptsize$\,\pm$2.3} & \cellcolor[rgb]{0.990,0.802,0.856}\textbf{40.0} & \cellcolor[rgb]{0.990,0.802,0.856}\textbf{0.252} & \cellcolor[rgb]{0.990,0.802,0.856}\textbf{369} & \cellcolor[rgb]{0.990,0.802,0.856}\textbf{543} & \cellcolor[rgb]{0.990,0.802,0.856}\textbf{0.370} \\
\midrule
\harnessicon{\iconopenai}Codex CLI & gpt-6-luna & \cellcolor[rgb]{0.984,0.992,1.000}6.39{\scriptsize$\,\pm$0.53} & \cellcolor[rgb]{0.984,0.992,1.000}45.3{\scriptsize$\,\pm$3.2} & \cellcolor[rgb]{0.984,0.992,1.000}40.3{\scriptsize$\,\pm$1.5} & \cellcolor[rgb]{0.984,0.992,1.000}32.7{\scriptsize$\,\pm$1.4} & \cellcolor[rgb]{0.990,0.802,0.856}\textbf{57} & \cellcolor[rgb]{0.999,0.984,0.988}0.03 & \cellcolor[rgb]{0.999,0.984,0.988}2083 & \cellcolor[rgb]{0.999,0.984,0.988}5168 & \cellcolor[rgb]{0.999,0.984,0.988}0.074 \\
\hdashline
\harnessicon{}\textbf{NavRSI (ours)} & gpt-6-luna & \cellcolor[rgb]{0.809,0.903,0.996}\textbf{5.16}{\scriptsize$\,\pm$0.74} & \cellcolor[rgb]{0.809,0.903,0.996}\textbf{57.3}{\scriptsize$\,\pm$8.4} & \cellcolor[rgb]{0.809,0.903,0.996}\textbf{51.3}{\scriptsize$\,\pm$6.7} & \cellcolor[rgb]{0.809,0.903,0.996}\textbf{41.9}{\scriptsize$\,\pm$4.2} & \cellcolor[rgb]{0.992,0.840,0.884}60.8 & \cellcolor[rgb]{0.990,0.802,0.856}\textbf{0.018} & \cellcolor[rgb]{0.990,0.802,0.856}\textbf{1038} & \cellcolor[rgb]{0.990,0.802,0.856}\textbf{2023} & \cellcolor[rgb]{0.990,0.802,0.856}\textbf{0.035} \\
\midrule
\harnessicon{\iconopenai}Codex CLI & gpt-6.1\kern-0.08em-terra & \cellcolor[rgb]{0.889,0.943,0.998}3.69{\scriptsize$\,\pm$0.19} & \cellcolor[rgb]{0.922,0.960,0.998}77.3{\scriptsize$\,\pm$2.3} & \cellcolor[rgb]{0.910,0.954,0.998}72.3{\scriptsize$\,\pm$0.6} & \cellcolor[rgb]{0.899,0.949,0.998}59.0{\scriptsize$\,\pm$0.8} & \cellcolor[rgb]{0.995,0.895,0.924}29.4 & \cellcolor[rgb]{0.999,0.984,0.988}0.061 & \cellcolor[rgb]{0.999,0.984,0.988}587 & \cellcolor[rgb]{0.999,0.984,0.988}811 & \cellcolor[rgb]{0.999,0.984,0.988}0.084 \\
\hdashline
\harnessicon{}\textbf{NavRSI (ours)} & gpt-6.1\kern-0.08em-terra & \cellcolor[rgb]{0.809,0.903,0.996}\textbf{3.55}{\scriptsize$\,\pm$0.29} & \cellcolor[rgb]{0.809,0.903,0.996}\textbf{82.3}{\scriptsize$\,\pm$1.5} & \cellcolor[rgb]{0.809,0.903,0.996}\textbf{76.3}{\scriptsize$\,\pm$1.5} & \cellcolor[rgb]{0.809,0.903,0.996}\textbf{61.8}{\scriptsize$\,\pm$0.7} & \cellcolor[rgb]{0.990,0.802,0.856}\textbf{24.9} & \cellcolor[rgb]{0.990,0.802,0.856}\textbf{0.021} & \cellcolor[rgb]{0.990,0.802,0.856}\textbf{161} & \cellcolor[rgb]{0.990,0.802,0.856}\textbf{211} & \cellcolor[rgb]{0.990,0.802,0.856}\textbf{0.027} \\
\midrule
\harnessicon{\iconopenai}Codex CLI & gpt-6.1\kern-0.08em-sol & \cellcolor[rgb]{0.944,0.972,0.999}3.44{\scriptsize$\,\pm$0.22} & \cellcolor[rgb]{0.885,0.941,0.997}78.3{\scriptsize$\,\pm$1.5} & \cellcolor[rgb]{0.881,0.939,0.997}75.0{\scriptsize$\,\pm$1.0} & \cellcolor[rgb]{0.920,0.959,0.998}62.3{\scriptsize$\,\pm$2.3} & \cellcolor[rgb]{0.993,0.861,0.899}30 & \cellcolor[rgb]{0.999,0.984,0.988}0.14 & \cellcolor[rgb]{0.999,0.984,0.988}673 & \cellcolor[rgb]{0.999,0.984,0.988}897 & \cellcolor[rgb]{0.999,0.984,0.988}0.187 \\
\hdashline
\harnessicon{}\textbf{NavRSI (ours)} & gpt-6.1\kern-0.08em-sol & \cellcolor[rgb]{0.809,0.903,0.996}\textbf{3.19}{\scriptsize$\,\pm$0.61} & \cellcolor[rgb]{0.809,0.903,0.996}\textbf{81.3}{\scriptsize$\,\pm$4.0} & \cellcolor[rgb]{0.809,0.903,0.996}\textbf{77.7}{\scriptsize$\,\pm$3.5} & \cellcolor[rgb]{0.809,0.903,0.996}\textbf{66.2}{\scriptsize$\,\pm$2.1} & \cellcolor[rgb]{0.990,0.802,0.856}\textbf{27.1} & \cellcolor[rgb]{0.990,0.802,0.856}\textbf{0.046} & \cellcolor[rgb]{0.990,0.802,0.856}\textbf{177} & \cellcolor[rgb]{0.990,0.802,0.856}\textbf{228} & \cellcolor[rgb]{0.990,0.802,0.856}\textbf{0.059} \\
\midrule
\harnessicon{\iconopenai}Codex CLI & gpt-6-astra & \cellcolor[rgb]{0.984,0.992,1.000}3.71{\scriptsize$\,\pm$0.50} & \cellcolor[rgb]{0.899,0.948,0.998}78.3{\scriptsize$\,\pm$4.2} & \cellcolor[rgb]{0.879,0.938,0.997}75.7{\scriptsize$\,\pm$4.7} & \cellcolor[rgb]{0.856,0.926,0.997}66.0{\scriptsize$\,\pm$3.5} & \cellcolor[rgb]{0.994,0.881,0.913}28 & \cellcolor[rgb]{0.999,0.984,0.988}0.90 & \cellcolor[rgb]{0.999,0.984,0.988}580 & \cellcolor[rgb]{0.999,0.984,0.988}766 & \cellcolor[rgb]{0.999,0.984,0.988}1.189 \\
\hdashline
\harnessicon{}\textbf{NavRSI (ours)} & gpt-6-astra & \cellcolor[rgb]{0.809,0.903,0.996}\textbf{3.13}{\scriptsize$\,\pm$0.29} & \cellcolor[rgb]{0.809,0.903,0.996}\textbf{82.0}{\scriptsize$\,\pm$1.7} & \cellcolor[rgb]{0.809,0.903,0.996}\textbf{78.3}{\scriptsize$\,\pm$1.5} & \cellcolor[rgb]{0.809,0.903,0.996}\textbf{67.4}{\scriptsize$\,\pm$0.6} & \cellcolor[rgb]{0.990,0.802,0.856}\textbf{24.4} & \cellcolor[rgb]{0.990,0.802,0.856}\textbf{0.272} & \cellcolor[rgb]{0.990,0.802,0.856}\textbf{153} & \cellcolor[rgb]{0.990,0.802,0.856}\textbf{195} & \cellcolor[rgb]{0.990,0.802,0.856}\textbf{0.347} \\
\midrule
\harnessicon{\iconopenai}Codex CLI & gpt-6.2-sol & \cellcolor[rgb]{0.984,0.992,1.000}3.00{\scriptsize$\,\pm$0.02} & \cellcolor[rgb]{0.931,0.965,0.998}88.0{\scriptsize$\,\pm$2.0} & \cellcolor[rgb]{0.945,0.972,0.999}82.0{\scriptsize$\,\pm$2.0} & \cellcolor[rgb]{0.966,0.983,0.999}70.9{\scriptsize$\,\pm$2.4} & \cellcolor[rgb]{0.995,0.895,0.923}26.5 & \cellcolor[rgb]{0.990,0.802,0.856}\textbf{0.097} & \cellcolor[rgb]{0.999,0.984,0.988}323 & \cellcolor[rgb]{0.999,0.984,0.988}394 & \cellcolor[rgb]{0.992,0.836,0.881}0.118 \\
\hdashline
\harnessicon{}\textbf{NavRSI (ours)} & gpt-6.2-sol & \cellcolor[rgb]{0.809,0.903,0.996}\textbf{2.18}{\scriptsize$\,\pm$0.64} & \cellcolor[rgb]{0.809,0.903,0.996}\textbf{94.3}{\scriptsize$\,\pm$2.1} & \cellcolor[rgb]{0.809,0.903,0.996}\textbf{89.0}{\scriptsize$\,\pm$1.7} & \cellcolor[rgb]{0.809,0.903,0.996}\textbf{78.9}{\scriptsize$\,\pm$0.5} & \cellcolor[rgb]{0.990,0.802,0.856}\textbf{22.5} & \cellcolor[rgb]{0.991,0.817,0.867}0.099 & \cellcolor[rgb]{0.990,0.802,0.856}\textbf{131} & \cellcolor[rgb]{0.990,0.802,0.856}\textbf{148} & \cellcolor[rgb]{0.990,0.802,0.856}\textbf{0.112} \\
\midrule
\harnessicon{\iconopenai}Codex CLI & gpt-6.2-astra & \cellcolor[rgb]{0.809,0.903,0.996}\textbf{1.80}{\scriptsize$\,\pm$0.18} & \cellcolor[rgb]{0.863,0.930,0.997}90.7{\scriptsize$\,\pm$1.2} & \cellcolor[rgb]{0.857,0.927,0.997}89.7{\scriptsize$\,\pm$1.2} & \cellcolor[rgb]{0.871,0.934,0.997}76.1{\scriptsize$\,\pm$1.1} & \cellcolor[rgb]{0.996,0.910,0.934}20.1 & \cellcolor[rgb]{0.997,0.933,0.951}0.241 & \cellcolor[rgb]{0.999,0.984,0.988}212 & \cellcolor[rgb]{0.999,0.984,0.988}237 & \cellcolor[rgb]{0.997,0.941,0.957}0.268 \\
\hdashline
\harnessicon{}\textbf{NavRSI (ours)} & gpt-6.2-astra & \cellcolor[rgb]{0.931,0.965,0.998}1.92{\scriptsize$\,\pm$0.31} & \cellcolor[rgb]{0.809,0.903,0.996}\textbf{93.0}{\scriptsize$\,\pm$1.0} & \cellcolor[rgb]{0.809,0.903,0.996}\textbf{91.7}{\scriptsize$\,\pm$1.2} & \cellcolor[rgb]{0.809,0.903,0.996}\textbf{78.3}{\scriptsize$\,\pm$0.5} & \cellcolor[rgb]{0.990,0.802,0.856}\textbf{16.5} & \cellcolor[rgb]{0.990,0.802,0.856}\textbf{0.188} & \cellcolor[rgb]{0.990,0.802,0.856}\textbf{77} & \cellcolor[rgb]{0.990,0.802,0.856}\textbf{84} & \cellcolor[rgb]{0.990,0.802,0.856}\textbf{0.205} \\
\midrule
\harnessicon{\icongemini}Antigravity & gemini-3.8-flash & \cellcolor[rgb]{0.984,0.992,1.000}3.40{\scriptsize$\,\pm$0.50} & \cellcolor[rgb]{0.920,0.959,0.998}79.7{\scriptsize$\,\pm$1.5} & \cellcolor[rgb]{0.948,0.974,0.999}74.7{\scriptsize$\,\pm$0.6} & \cellcolor[rgb]{0.940,0.969,0.999}63.5{\scriptsize$\,\pm$0.2} & \cellcolor[rgb]{0.991,0.816,0.866}50.8 & \cellcolor[rgb]{0.991,0.813,0.864}0.304 & \cellcolor[rgb]{0.991,0.827,0.874}1968 & \cellcolor[rgb]{0.994,0.875,0.909}2635 & \cellcolor[rgb]{0.993,0.862,0.899}0.408 \\
\hdashline
\harnessicon{}\textbf{NavRSI (ours)} & gemini-3.8-flash & \cellcolor[rgb]{0.809,0.903,0.996}\textbf{2.89}{\scriptsize$\,\pm$0.54} & \cellcolor[rgb]{0.809,0.903,0.996}\textbf{84.7}{\scriptsize$\,\pm$2.5} & \cellcolor[rgb]{0.809,0.903,0.996}\textbf{81.3}{\scriptsize$\,\pm$0.6} & \cellcolor[rgb]{0.809,0.903,0.996}\textbf{68.6}{\scriptsize$\,\pm$1.3} & \cellcolor[rgb]{0.990,0.802,0.856}\textbf{49.7} & \cellcolor[rgb]{0.990,0.802,0.856}\textbf{0.299} & \cellcolor[rgb]{0.990,0.802,0.856}\textbf{1886} & \cellcolor[rgb]{0.990,0.802,0.856}\textbf{2319} & \cellcolor[rgb]{0.990,0.802,0.856}\textbf{0.368} \\
\midrule
\harnessicon{\icongemini}Antigravity & gemini-4-hydrogen & \cellcolor[rgb]{0.977,0.988,0.999}2.63{\scriptsize$\,\pm$0.41} & \cellcolor[rgb]{0.833,0.915,0.996}94.0{\scriptsize$\,\pm$1.0} & \cellcolor[rgb]{0.835,0.915,0.996}88.0{\scriptsize$\,\pm$2.6} & \cellcolor[rgb]{0.829,0.913,0.996}73.2{\scriptsize$\,\pm$0.9} & \cellcolor[rgb]{0.994,0.872,0.907}35.2 & \cellcolor[rgb]{0.993,0.852,0.892}0.286 & \cellcolor[rgb]{0.999,0.983,0.987}1018 & \cellcolor[rgb]{0.999,0.983,0.988}1157 & \cellcolor[rgb]{0.993,0.859,0.897}0.326 \\
\hdashline
\harnessicon{}\textbf{NavRSI (ours)} & gemini-4-hydrogen & \cellcolor[rgb]{0.809,0.903,0.996}\textbf{2.35}{\scriptsize$\,\pm$0.47} & \cellcolor[rgb]{0.809,0.903,0.996}\textbf{95.0}{\scriptsize$\,\pm$1.7} & \cellcolor[rgb]{0.809,0.903,0.996}\textbf{89.0}{\scriptsize$\,\pm$3.0} & \cellcolor[rgb]{0.809,0.903,0.996}\textbf{73.8}{\scriptsize$\,\pm$3.4} & \cellcolor[rgb]{0.990,0.802,0.856}\textbf{31.1} & \cellcolor[rgb]{0.990,0.802,0.856}\textbf{0.263} & \cellcolor[rgb]{0.990,0.802,0.856}\textbf{654} & \cellcolor[rgb]{0.990,0.802,0.856}\textbf{734} & \cellcolor[rgb]{0.990,0.802,0.856}\textbf{0.295} \\
\midrule
\harnessicon{\iconqwen}Qwen Code & qwen3.8-flash & \cellcolor[rgb]{0.978,0.989,1.000}4.71{\scriptsize$\,\pm$0.24} & \cellcolor[rgb]{0.809,0.903,0.996}\textbf{70.0}{\scriptsize$\,\pm$1.0} & \cellcolor[rgb]{0.833,0.915,0.996}61.7{\scriptsize$\,\pm$3.8} & \cellcolor[rgb]{0.809,0.903,0.996}\textbf{49.9}{\scriptsize$\,\pm$2.3} & \cellcolor[rgb]{0.994,0.888,0.918}75.1 & \cellcolor[rgb]{0.999,0.984,0.988}0.077 & \cellcolor[rgb]{0.999,0.983,0.988}2555 & \cellcolor[rgb]{0.999,0.983,0.988}4144 & \cellcolor[rgb]{0.999,0.984,0.988}0.125 \\
\hdashline
\harnessicon{}\textbf{NavRSI (ours)} & qwen3.8-flash & \cellcolor[rgb]{0.809,0.903,0.996}\textbf{4.19}{\scriptsize$\,\pm$0.56} & \cellcolor[rgb]{0.860,0.929,0.997}68.3{\scriptsize$\,\pm$8.1} & \cellcolor[rgb]{0.809,0.903,0.996}\textbf{62.3}{\scriptsize$\,\pm$6.8} & \cellcolor[rgb]{0.825,0.911,0.996}49.6{\scriptsize$\,\pm$5.2} & \cellcolor[rgb]{0.990,0.802,0.856}\textbf{64.5} & \cellcolor[rgb]{0.990,0.802,0.856}\textbf{0.039} & \cellcolor[rgb]{0.990,0.802,0.856}\textbf{1629} & \cellcolor[rgb]{0.990,0.802,0.856}\textbf{2613} & \cellcolor[rgb]{0.990,0.802,0.856}\textbf{0.063} \\
\midrule
\harnessicon{\iconqwen}Qwen Code & qwen3.7-flash & \cellcolor[rgb]{0.984,0.992,1.000}7.99{\scriptsize$\,\pm$0.28} & \cellcolor[rgb]{0.984,0.992,1.000}54.3{\scriptsize$\,\pm$2.9} & \cellcolor[rgb]{0.984,0.992,1.000}25.3{\scriptsize$\,\pm$1.2} & \cellcolor[rgb]{0.984,0.992,1.000}15.8{\scriptsize$\,\pm$1.2} & \cellcolor[rgb]{0.990,0.804,0.857}65.1 & \cellcolor[rgb]{0.997,0.942,0.958}0.006 & \cellcolor[rgb]{0.992,0.838,0.882}834 & \cellcolor[rgb]{0.999,0.978,0.984}3291 & \cellcolor[rgb]{0.999,0.984,0.988}0.022 \\
\hdashline
\harnessicon{}\textbf{NavRSI (ours)} & qwen3.7-flash & \cellcolor[rgb]{0.809,0.903,0.996}\textbf{6.70}{\scriptsize$\,\pm$0.79} & \cellcolor[rgb]{0.809,0.903,0.996}\textbf{66.3}{\scriptsize$\,\pm$7.5} & \cellcolor[rgb]{0.809,0.903,0.996}\textbf{35.7}{\scriptsize$\,\pm$4.2} & \cellcolor[rgb]{0.809,0.903,0.996}\textbf{22.3}{\scriptsize$\,\pm$0.7} & \cellcolor[rgb]{0.990,0.802,0.856}\textbf{64.9} & \cellcolor[rgb]{0.990,0.802,0.856}\textbf{0.004} & \cellcolor[rgb]{0.990,0.802,0.856}\textbf{784} & \cellcolor[rgb]{0.990,0.802,0.856}\textbf{2199} & \cellcolor[rgb]{0.990,0.802,0.856}\textbf{0.012} \\
\midrule
\harnessicon{\iconqwen}Qwen Code & qwen3.8-27b & \cellcolor[rgb]{0.958,0.978,0.999}5.44{\scriptsize$\,\pm$0.41} & \cellcolor[rgb]{0.907,0.953,0.998}56.3{\scriptsize$\,\pm$5.0} & \cellcolor[rgb]{0.962,0.980,0.999}50.0{\scriptsize$\,\pm$3.5} & \cellcolor[rgb]{0.906,0.952,0.998}42.9{\scriptsize$\,\pm$3.6} & \cellcolor[rgb]{0.990,0.802,0.856}\textbf{61.5} & \cellcolor[rgb]{0.991,0.829,0.875}1.033 & \cellcolor[rgb]{0.991,0.810,0.862}1884 & \cellcolor[rgb]{0.993,0.868,0.904}3769 & \cellcolor[rgb]{0.994,0.885,0.916}2.065 \\
\hdashline
\harnessicon{}\textbf{NavRSI (ours)} & qwen3.8-27b & \cellcolor[rgb]{0.809,0.903,0.996}\textbf{4.99}{\scriptsize$\,\pm$0.56} & \cellcolor[rgb]{0.809,0.903,0.996}\textbf{59.3}{\scriptsize$\,\pm$3.8} & \cellcolor[rgb]{0.809,0.903,0.996}\textbf{55.3}{\scriptsize$\,\pm$5.0} & \cellcolor[rgb]{0.809,0.903,0.996}\textbf{45.1}{\scriptsize$\,\pm$3.1} & \cellcolor[rgb]{0.991,0.821,0.870}63.5 & \cellcolor[rgb]{0.990,0.802,0.856}\textbf{0.988} & \cellcolor[rgb]{0.990,0.802,0.856}\textbf{1860} & \cellcolor[rgb]{0.990,0.802,0.856}\textbf{3361} & \cellcolor[rgb]{0.990,0.802,0.856}\textbf{1.785} \\
\specialrule{1pt}{0pt}{0pt}
\end{tabular}%
}
\end{table*}

\begin{table*}[t]
\caption{\textbf{In-distribution held-out set of R2R-CE val-unseen across LLM backbones} (537 episodes in total, including 100 episodes in R2R-CE-100~\cite{qiao2025opennav}). With every backbone, NavRSI attains the best SR with the fewest tokens from a single RGB camera and no training.}
\label{tab:r2r_heldout_llms}
\centering
\footnotesize
\providecommand{\flame}{\,\raisebox{-0.14em}{\includegraphics[height=0.95em]{figures/icons/flame.pdf}}}
\setlength{\tabcolsep}{3.4pt}
\setlength{\aboverulesep}{0pt}
\setlength{\belowrulesep}{0pt}
\renewcommand{\arraystretch}{1.2}
\setlength{\dashlinedash}{2pt}
\setlength{\dashlinegap}{1.5pt}
\resizebox{\textwidth}{!}{%
\begin{tabular}{lcccccccccc}
\specialrule{1pt}{0pt}{0pt}
\rowcolor[gray]{0.9} & \multicolumn{5}{c}{Navigation} & \multicolumn{5}{c}{Resource use} \\
\noalign{\vskip-\cmidrulewidth}\cmidrule(lr){2-6}\cmidrule(lr){7-11}
\rowcolor[gray]{0.9}\multirow{-2}{*}{Method} & NE$\downarrow$ & OSR$\uparrow$ & SR$\uparrow$ & SPL$\uparrow$ & nDTW$\uparrow$ & GPU Mem. (GB)$\downarrow$ & Cost (\$)$\downarrow$ & Tokens (k)$\downarrow$ & Tokens/SR (k)$\downarrow$ & Cost/SR (\$)$\downarrow$ \\
\specialrule{0.5pt}{0pt}{1.2pt}\specialrule{0.5pt}{0pt}{0pt}
\multicolumn{11}{l}{\cellcolor[gray]{0.96}\textit{LLM backbone: GPT-6.2-Astra}} \\
\hdashline
MIP~\cite{zhou2026embodied} & \cellcolor[rgb]{0.984,0.992,1.000}2.02{\scriptsize$\,\pm$0.04} & \cellcolor[rgb]{0.984,0.992,1.000}89.8{\scriptsize$\,\pm$0.2} & \cellcolor[rgb]{0.967,0.983,1.000}86.8{\scriptsize$\,\pm$0.4} & \cellcolor[rgb]{0.984,0.992,1.000}73.6{\scriptsize$\,\pm$0.5} & \cellcolor[rgb]{0.905,0.952,0.998}76.3{\scriptsize$\,\pm$0.4} & \cellcolor[rgb]{0.990,0.802,0.856}\textbf{0.93} & \cellcolor[rgb]{0.999,0.984,0.988}0.231 & \cellcolor[rgb]{0.999,0.984,0.988}204 & \cellcolor[rgb]{0.999,0.984,0.988}235 & \cellcolor[rgb]{0.999,0.984,0.988}0.266 \\
\hdashline
\textsc{NavGPT-3}~\cite{zhou2026navgpt3} + 8B VLA\flame & \cellcolor[rgb]{0.919,0.959,0.999}1.88{\scriptsize$\,\pm$0.02} & \cellcolor[rgb]{0.969,0.984,1.000}90.1{\scriptsize$\,\pm$0.2} & \cellcolor[rgb]{0.984,0.992,1.000}86.5{\scriptsize$\,\pm$0.5} & \cellcolor[rgb]{0.903,0.951,0.998}75.3{\scriptsize$\,\pm$0.7} & \cellcolor[rgb]{0.984,0.992,1.000}73.1{\scriptsize$\,\pm$0.2} & \cellcolor[rgb]{0.999,0.984,0.988}18.97 & \cellcolor[rgb]{0.990,0.802,0.856}\textbf{0.205} & \cellcolor[rgb]{0.991,0.830,0.876}95 & \cellcolor[rgb]{0.992,0.834,0.879}110 & \cellcolor[rgb]{0.990,0.802,0.856}\textbf{0.237} \\
\hdashline
\textbf{NavRSI (ours)} & \cellcolor[rgb]{0.809,0.903,0.996}\textbf{1.64}{\scriptsize$\,\pm$0.07} & \cellcolor[rgb]{0.809,0.903,0.996}\textbf{92.7}{\scriptsize$\,\pm$0.5} & \cellcolor[rgb]{0.809,0.903,0.996}\textbf{90.3}{\scriptsize$\,\pm$0.5} & \cellcolor[rgb]{0.809,0.903,0.996}\textbf{77.2}{\scriptsize$\,\pm$0.9} & \cellcolor[rgb]{0.809,0.903,0.996}\textbf{80.0}{\scriptsize$\,\pm$1.0} & \cellcolor[rgb]{0.990,0.802,0.856}\textbf{0.93} & \cellcolor[rgb]{0.998,0.972,0.979}0.229 & \cellcolor[rgb]{0.990,0.802,0.856}\textbf{76} & \cellcolor[rgb]{0.990,0.802,0.856}\textbf{84} & \cellcolor[rgb]{0.995,0.907,0.932}0.254 \\
\specialrule{0.5pt}{0pt}{0pt}
\multicolumn{11}{l}{\cellcolor[gray]{0.96}\textit{LLM backbone: GPT-6.2-Sol}} \\
\hdashline
MIP~\cite{zhou2026embodied} & \cellcolor[rgb]{0.984,0.992,1.000}2.42{\scriptsize$\,\pm$0.24} & \cellcolor[rgb]{0.984,0.992,1.000}88.8{\scriptsize$\,\pm$0.5} & \cellcolor[rgb]{0.984,0.992,1.000}83.9{\scriptsize$\,\pm$1.0} & \cellcolor[rgb]{0.984,0.992,1.000}71.2{\scriptsize$\,\pm$0.9} & \cellcolor[rgb]{0.972,0.986,1.000}73.8{\scriptsize$\,\pm$0.6} & \cellcolor[rgb]{0.990,0.802,0.856}\textbf{0.93} & \cellcolor[rgb]{0.999,0.983,0.987}0.100 & \cellcolor[rgb]{0.999,0.984,0.988}317 & \cellcolor[rgb]{0.999,0.984,0.988}378 & \cellcolor[rgb]{0.999,0.984,0.988}0.119 \\
\hdashline
\textsc{NavGPT-3}~\cite{zhou2026navgpt3} + 8B VLA\flame & \cellcolor[rgb]{0.809,0.903,0.996}\textbf{1.93}{\scriptsize$\,\pm$0.10} & \cellcolor[rgb]{0.926,0.962,0.999}89.8{\scriptsize$\,\pm$0.5} & \cellcolor[rgb]{0.923,0.961,0.999}85.3{\scriptsize$\,\pm$0.0} & \cellcolor[rgb]{0.866,0.932,0.997}74.6{\scriptsize$\,\pm$0.3} & \cellcolor[rgb]{0.984,0.992,1.000}73.5{\scriptsize$\,\pm$0.7} & \cellcolor[rgb]{0.999,0.984,0.988}18.97 & \cellcolor[rgb]{0.990,0.802,0.856}\textbf{0.073} & \cellcolor[rgb]{0.990,0.811,0.863}143 & \cellcolor[rgb]{0.991,0.814,0.865}168 & \cellcolor[rgb]{0.990,0.802,0.856}\textbf{0.086} \\
\hdashline
\textbf{NavRSI (ours)} & \cellcolor[rgb]{0.846,0.922,0.997}2.03{\scriptsize$\,\pm$0.11} & \cellcolor[rgb]{0.809,0.903,0.996}\textbf{91.8}{\scriptsize$\,\pm$0.6} & \cellcolor[rgb]{0.809,0.903,0.996}\textbf{87.8}{\scriptsize$\,\pm$0.4} & \cellcolor[rgb]{0.809,0.903,0.996}\textbf{76.2}{\scriptsize$\,\pm$0.5} & \cellcolor[rgb]{0.809,0.903,0.996}\textbf{78.8}{\scriptsize$\,\pm$0.6} & \cellcolor[rgb]{0.990,0.802,0.856}\textbf{0.93} & \cellcolor[rgb]{0.999,0.984,0.988}0.100 & \cellcolor[rgb]{0.990,0.802,0.856}\textbf{134} & \cellcolor[rgb]{0.990,0.802,0.856}\textbf{152} & \cellcolor[rgb]{0.998,0.955,0.967}0.114 \\
\specialrule{0.5pt}{0pt}{0pt}
\multicolumn{11}{l}{\cellcolor[gray]{0.96}\textit{LLM backbone: GPT-6-Astra}} \\
\hdashline
MIP~\cite{zhou2026embodied} & \cellcolor[rgb]{0.984,0.992,1.000}2.70{\scriptsize$\,\pm$0.10} & \cellcolor[rgb]{0.984,0.992,1.000}85.5{\scriptsize$\,\pm$0.4} & \cellcolor[rgb]{0.961,0.980,0.999}81.2{\scriptsize$\,\pm$0.2} & \cellcolor[rgb]{0.984,0.992,1.000}69.9{\scriptsize$\,\pm$0.2} & \cellcolor[rgb]{0.984,0.992,1.000}72.5{\scriptsize$\,\pm$0.2} & \cellcolor[rgb]{0.990,0.802,0.856}\textbf{0.93} & \cellcolor[rgb]{0.999,0.984,0.988}0.868 & \cellcolor[rgb]{0.999,0.984,0.988}514 & \cellcolor[rgb]{0.999,0.984,0.988}634 & \cellcolor[rgb]{0.999,0.984,0.988}1.069 \\
\hdashline
\textsc{NavGPT-3}~\cite{zhou2026navgpt3} + 8B VLA\flame & \cellcolor[rgb]{0.809,0.903,0.996}\textbf{2.15}{\scriptsize$\,\pm$0.07} & \cellcolor[rgb]{0.809,0.903,0.996}\textbf{87.8}{\scriptsize$\,\pm$0.2} & \cellcolor[rgb]{0.984,0.992,1.000}80.9{\scriptsize$\,\pm$0.3} & \cellcolor[rgb]{0.811,0.904,0.996}\textbf{72.1}{\scriptsize$\,\pm$0.4} & \cellcolor[rgb]{0.931,0.965,0.999}73.2{\scriptsize$\,\pm$0.4} & \cellcolor[rgb]{0.999,0.984,0.988}18.97 & \cellcolor[rgb]{0.992,0.840,0.883}0.587 & \cellcolor[rgb]{0.992,0.853,0.893}246 & \cellcolor[rgb]{0.993,0.854,0.894}304 & \cellcolor[rgb]{0.992,0.845,0.887}0.725 \\
\hdashline
\textbf{NavRSI (ours)} & \cellcolor[rgb]{0.903,0.951,0.998}2.45{\scriptsize$\,\pm$0.09} & \cellcolor[rgb]{0.897,0.948,0.998}86.7{\scriptsize$\,\pm$1.1} & \cellcolor[rgb]{0.809,0.903,0.996}\textbf{82.8}{\scriptsize$\,\pm$0.6} & \cellcolor[rgb]{0.809,0.903,0.996}\textbf{72.1}{\scriptsize$\,\pm$0.8} & \cellcolor[rgb]{0.809,0.903,0.996}\textbf{74.9}{\scriptsize$\,\pm$0.7} & \cellcolor[rgb]{0.990,0.802,0.856}\textbf{0.93} & \cellcolor[rgb]{0.990,0.802,0.856}\textbf{0.514} & \cellcolor[rgb]{0.990,0.802,0.856}\textbf{143} & \cellcolor[rgb]{0.990,0.802,0.856}\textbf{173} & \cellcolor[rgb]{0.990,0.802,0.856}\textbf{0.620} \\
\specialrule{1pt}{0pt}{0pt}
\end{tabular}%
}
\end{table*}

\begin{table*}[t]
\makeatletter\renewenvironment{table}[1][]{\def\@captype{table}}{}\makeatother
\begin{minipage}[t]{\columnwidth}
\begin{table}[t]
\caption{\textbf{Comparison in the real world} (average of 4 runs).}
\label{tab:realworld}
\centering
\footnotesize
\setlength{\aboverulesep}{0pt}
\setlength{\belowrulesep}{0pt}
\renewcommand{\arraystretch}{1.7}%
\setlength{\dashlinedash}{2pt}
\setlength{\dashlinegap}{1.5pt}
\newcommand{\rwbody}{%
\specialrule{1pt}{0pt}{0pt}
\rowcolor[gray]{0.9}Method & NE$\downarrow$ & OSR$\uparrow$ & SR$\uparrow$ \\
\specialrule{0.5pt}{0pt}{0pt}
Open-Nav~\cite{qiao2025opennav} (GPT-6-Astra) & 7.2 & 24 & 16 \\
\textsc{NavGPT-3}~\cite{zhou2026navgpt3} (GPT-6-Astra) & 6.9 & 44 & 36 \\
\hdashline
\textbf{NavRSI (ours)} (GPT-6-Astra) & \textbf{5.3} & \textbf{52} & \textbf{48} \\
\specialrule{1pt}{0pt}{0pt}}
\newlength{\rwnat}
\setlength{\tabcolsep}{0pt}
\settowidth{\rwnat}{\begin{tabular}{lccc}\rwbody\end{tabular}}
\setlength{\tabcolsep}{\dimexpr(\columnwidth-\rwnat)/8\relax}
\begin{tabular}{lccc}
\rwbody
\end{tabular}
\end{table}

\end{minipage}\hfill
\begin{minipage}[t]{\columnwidth}
\begin{table}[t]
\caption{\textbf{What experience accumulation brings} (R$k$: round $k$).}
\label{tab:abl_exp}
\centering
\footnotesize
\setlength{\tabcolsep}{3.4pt}
\setlength{\aboverulesep}{0pt}
\setlength{\belowrulesep}{0pt}
\renewcommand{\arraystretch}{1.15}
\begin{tabular}{lcccccc}
\specialrule{1pt}{0pt}{0pt}
\rowcolor[gray]{0.9} & Size & \multicolumn{4}{c}{Repeated rewrites (\%)$\downarrow$} & Input \\
\noalign{\vskip-\cmidrulewidth}\cmidrule(lr){3-6}
\rowcolor[gray]{0.9}\multirow{-2}{*}{The designer may read} & (words)$\downarrow$ & R4 & R6 & R8 & All & (M tok.)$\downarrow$ \\
\specialrule{0.5pt}{0pt}{0pt}
\multicolumn{7}{l}{\rqtag{6}\hspace{0.4em}\textit{What does experience accumulation bring?}} \\
Current harness only & -- & 8.3 & 50.0 & 100.0 & 52.8 & 1.01 \\
+ full record & 13.1M & 0.0 & 0.0 & 0.0 & 0.0 & 1.16 \\
\textbf{+ experience $E_k$ only} & \textbf{2.2k} & \textbf{0.0} & \textbf{0.0} & \textbf{0.0} & \textbf{0.0} & \textbf{0.91} \\
\specialrule{1pt}{0pt}{0pt}
\end{tabular}
\end{table}

\end{minipage}
\vspace{-6pt}%
\end{table*}

\begingroup
\let\evoTable\table
\def\table[#1]{\evoTable[#1]{\centering\includegraphics[width=\linewidth]{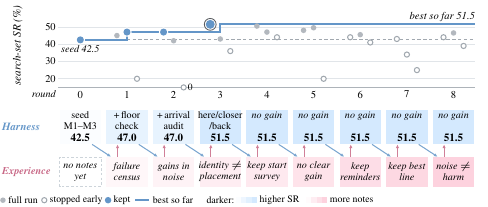}%
\captionof{figure}{%
\textbf{Harness rewriting in RSI.} Top: search-set SR of all 24 rewrites. Bottom: each round's experience from the designer.}
\label{fig:evolution}\par}\vspace{0pt}}
\begin{table}[t]
\caption{\textbf{From Seed 0 to the final harness on R2R-CE-100~\cite{qiao2025opennav}.} RSI ran with gpt-6-luna only.
Mean $\pm$ s.d. over three runs.}
\label{tab:r2r_mip}
\centering
\footnotesize
\setlength{\tabcolsep}{2.4pt}
\setlength{\aboverulesep}{0pt}
\setlength{\belowrulesep}{0pt}
\renewcommand{\arraystretch}{1.2}
\setlength{\dashlinedash}{2pt}
\setlength{\dashlinegap}{1.5pt}
\resizebox{\columnwidth}{!}{%
\begin{tabular}{llccccc}
\specialrule{1pt}{0pt}{0pt}
\rowcolor[gray]{0.9} & & \multicolumn{2}{c}{Navigation} & \multicolumn{3}{c}{Resource use} \\
\noalign{\vskip-\cmidrulewidth}\cmidrule(lr){3-4}\cmidrule(lr){5-7}
\rowcolor[gray]{0.9}\multirow{-2}{*}{Harness} & \multirow{-2}{*}{LLMs} & SR$\uparrow$ & SPL$\uparrow$ & Turns$\downarrow$ & Cost (\$)$\downarrow$ & Tokens (k)$\downarrow$ \\
\specialrule{0.5pt}{0pt}{1.2pt}\specialrule{0.5pt}{0pt}{0pt}
Seed 0 & gpt-5.6-sol & \cellcolor[rgb]{0.983,0.991,1.000}58.7{\scriptsize$\,\pm$4.7} & \cellcolor[rgb]{0.984,0.992,1.000}43.7{\scriptsize$\,\pm$1.7} & \cellcolor[rgb]{0.994,0.885,0.916}46.3 & \cellcolor[rgb]{0.990,0.802,0.856}\textbf{0.251} & \cellcolor[rgb]{0.991,0.818,0.868}379 \\
\hdashline
\textbf{NavRSI (ours)} & gpt-5.6-sol & \cellcolor[rgb]{0.809,0.903,0.996}\textbf{68.0}{\scriptsize$\,\pm$4.0}\,{\tiny\textcolor[rgb]{0,0.50,0.16}{$\uparrow$15.9\%}} & \cellcolor[rgb]{0.809,0.903,0.996}\textbf{53.5}{\scriptsize$\,\pm$2.3}\,{\tiny\textcolor[rgb]{0,0.50,0.16}{$\uparrow$22.5\%}} & \cellcolor[rgb]{0.990,0.802,0.856}\textbf{40.0}\,{\tiny\textcolor[rgb]{0,0.50,0.16}{$\downarrow$13.6\%}} & \cellcolor[rgb]{0.990,0.804,0.857}0.252\,{\tiny\textcolor[rgb]{0.50,0.50,0.50}{$\uparrow$0.2\%}} & \cellcolor[rgb]{0.990,0.802,0.856}\textbf{369}\,{\tiny\textcolor[rgb]{0,0.50,0.16}{$\downarrow$2.6\%}} \\
\midrule
Seed 0 & gpt-6-luna & \cellcolor[rgb]{0.984,0.992,1.000}43.7{\scriptsize$\,\pm$5.1} & \cellcolor[rgb]{0.982,0.991,1.000}36.2{\scriptsize$\,\pm$4.7} & \cellcolor[rgb]{0.994,0.877,0.910}69.3 & \cellcolor[rgb]{0.990,0.802,0.856}\textbf{0.016} & \cellcolor[rgb]{0.992,0.844,0.886}1114 \\
\hdashline
\textbf{NavRSI (ours)} & gpt-6-luna & \cellcolor[rgb]{0.809,0.903,0.996}\textbf{51.3}{\scriptsize$\,\pm$6.7}\,{\tiny\textcolor[rgb]{0,0.50,0.16}{$\uparrow$17.4\%}} & \cellcolor[rgb]{0.809,0.903,0.996}\textbf{41.9}{\scriptsize$\,\pm$4.2}\,{\tiny\textcolor[rgb]{0,0.50,0.16}{$\uparrow$15.6\%}} & \cellcolor[rgb]{0.990,0.802,0.856}\textbf{60.8}\,{\tiny\textcolor[rgb]{0,0.50,0.16}{$\downarrow$12.3\%}} & \cellcolor[rgb]{0.993,0.860,0.898}0.018\,{\tiny\textcolor[rgb]{0.50,0.50,0.50}{$\uparrow$10.4\%}} & \cellcolor[rgb]{0.990,0.802,0.856}\textbf{1038}\,{\tiny\textcolor[rgb]{0,0.50,0.16}{$\downarrow$6.8\%}} \\
\midrule
Seed 0 & gpt-6.1\kern-0.08em-terra & \cellcolor[rgb]{0.896,0.947,0.998}73.0{\scriptsize$\,\pm$3.0} & \cellcolor[rgb]{0.829,0.913,0.996}61.3{\scriptsize$\,\pm$3.9} & \cellcolor[rgb]{0.992,0.833,0.878}26.2 & \cellcolor[rgb]{0.990,0.802,0.856}\textbf{0.016} & \cellcolor[rgb]{0.990,0.802,0.856}\textbf{131} \\
\hdashline
\textbf{NavRSI (ours)} & gpt-6.1\kern-0.08em-terra & \cellcolor[rgb]{0.809,0.903,0.996}\textbf{76.3}{\scriptsize$\,\pm$1.5}\,{\tiny\textcolor[rgb]{0,0.50,0.16}{$\uparrow$4.5\%}} & \cellcolor[rgb]{0.809,0.903,0.996}\textbf{61.8}{\scriptsize$\,\pm$0.7}\,{\tiny\textcolor[rgb]{0,0.50,0.16}{$\uparrow$0.9\%}} & \cellcolor[rgb]{0.990,0.802,0.856}\textbf{24.9}\,{\tiny\textcolor[rgb]{0,0.50,0.16}{$\downarrow$5.0\%}} & \cellcolor[rgb]{0.997,0.931,0.950}0.021\,{\tiny\textcolor[rgb]{0.50,0.50,0.50}{$\uparrow$27.8\%}} & \cellcolor[rgb]{0.996,0.915,0.938}161\,{\tiny\textcolor[rgb]{0.50,0.50,0.50}{$\uparrow$23.0\%}} \\
\midrule
Seed 0 & gpt-6.1\kern-0.08em-sol & \cellcolor[rgb]{0.939,0.969,0.999}72.0{\scriptsize$\,\pm$2.0} & \cellcolor[rgb]{0.953,0.976,0.999}60.5{\scriptsize$\,\pm$2.6} & \cellcolor[rgb]{0.994,0.870,0.905}30.5 & \cellcolor[rgb]{0.990,0.802,0.856}\textbf{0.039} & \cellcolor[rgb]{0.990,0.802,0.856}\textbf{168} \\
\hdashline
\textbf{NavRSI (ours)} & gpt-6.1\kern-0.08em-sol & \cellcolor[rgb]{0.809,0.903,0.996}\textbf{77.7}{\scriptsize$\,\pm$3.5}\,{\tiny\textcolor[rgb]{0,0.50,0.16}{$\uparrow$7.9\%}} & \cellcolor[rgb]{0.809,0.903,0.996}\textbf{66.2}{\scriptsize$\,\pm$2.1}\,{\tiny\textcolor[rgb]{0,0.50,0.16}{$\uparrow$9.5\%}} & \cellcolor[rgb]{0.990,0.802,0.856}\textbf{27.1}\,{\tiny\textcolor[rgb]{0,0.50,0.16}{$\downarrow$11.1\%}} & \cellcolor[rgb]{0.994,0.889,0.919}0.046\,{\tiny\textcolor[rgb]{0.50,0.50,0.50}{$\uparrow$16.6\%}} & \cellcolor[rgb]{0.992,0.832,0.878}177\,{\tiny\textcolor[rgb]{0.50,0.50,0.50}{$\uparrow$5.1\%}} \\
\midrule
Seed 0 & gpt-6-astra & \cellcolor[rgb]{0.909,0.953,0.998}74.3{\scriptsize$\,\pm$2.1} & \cellcolor[rgb]{0.919,0.959,0.998}63.5{\scriptsize$\,\pm$1.5} & \cellcolor[rgb]{0.991,0.829,0.875}25.5 & \cellcolor[rgb]{0.990,0.802,0.856}\textbf{0.203} & \cellcolor[rgb]{0.990,0.802,0.856}\textbf{118} \\
\hdashline
\textbf{NavRSI (ours)} & gpt-6-astra & \cellcolor[rgb]{0.809,0.903,0.996}\textbf{78.3}{\scriptsize$\,\pm$1.5}\,{\tiny\textcolor[rgb]{0,0.50,0.16}{$\uparrow$5.4\%}} & \cellcolor[rgb]{0.809,0.903,0.996}\textbf{67.4}{\scriptsize$\,\pm$0.6}\,{\tiny\textcolor[rgb]{0,0.50,0.16}{$\uparrow$6.2\%}} & \cellcolor[rgb]{0.990,0.802,0.856}\textbf{24.4}\,{\tiny\textcolor[rgb]{0,0.50,0.16}{$\downarrow$4.3\%}} & \cellcolor[rgb]{0.997,0.949,0.963}0.272\,{\tiny\textcolor[rgb]{0.50,0.50,0.50}{$\uparrow$33.7\%}} & \cellcolor[rgb]{0.997,0.937,0.954}153\,{\tiny\textcolor[rgb]{0.50,0.50,0.50}{$\uparrow$29.4\%}} \\
\specialrule{1pt}{0pt}{0pt}
\end{tabular}%
}
\end{table}

\endgroup

\begin{table}[t]
\caption{\textbf{Closed-loop rewriting on R2R-CE-100}~\cite{qiao2025opennav}. One-shot: written once by the designer from Seed 0, with no closed-loop feedback. Round $K$: best harness after round $K$ of RSI.}
\label{tab:abl_rounds}\label{tab:abl_oneshot}
\centering
\footnotesize
\setlength{\aboverulesep}{0pt}
\setlength{\belowrulesep}{0pt}
\newcommand{\rdbody}{%
\specialrule{1pt}{0pt}{0pt}
\rowcolor[gray]{0.9} Harness & Search set (SR) & gpt-6-luna (SR) & gpt-6.1-sol (SR) \\
\specialrule{0.5pt}{0pt}{0pt}
\multicolumn{4}{l}{\rqtag{5}\hspace{0.4em}\textit{Is closed-loop harness rewriting effective?}} \\
Seed 0 & -- & 43.7{\scriptsize$\,\pm$5.1} & 72.0{\scriptsize$\,\pm$2.0} \\
One-shot design A & -- & 34.3{\scriptsize$\,\pm$0.6} & 70.7{\scriptsize$\,\pm$1.5} \\
One-shot design B & -- & 40.7{\scriptsize$\,\pm$1.1} & 70.0{\scriptsize$\,\pm$1.7} \\
One-shot design C & -- & 40.7{\scriptsize$\,\pm$3.5} & 65.3{\scriptsize$\,\pm$3.8} \\
\midrule
RSI, round 0 & 42.5 & 45.7{\scriptsize$\,\pm$3.8} & 74.0{\scriptsize$\,\pm$1.7} \\
RSI, round 1 & 47.0 & 45.0{\scriptsize$\,\pm$4.0} & 73.3{\scriptsize$\,\pm$3.1} \\
RSI, round 2 & 47.0 & 47.3{\scriptsize$\,\pm$2.5} & 75.0{\scriptsize$\,\pm$2.0} \\
\textbf{RSI, rounds 3--8 (final)} & \textbf{51.5} & \textbf{51.3}{\scriptsize$\,\pm$6.7} & \textbf{77.7}{\scriptsize$\,\pm$3.5} \\
\specialrule{1pt}{0pt}{0pt}}
\newlength{\rdnat}
\setlength{\tabcolsep}{0pt}
\settowidth{\rdnat}{\begin{tabular}{lccc}\rdbody\end{tabular}}
\setlength{\tabcolsep}{\dimexpr(\columnwidth-\rdnat)/8\relax}
\begin{tabular}{lccc}
\rdbody
\end{tabular}
\end{table}

\begin{table*}[t]
\caption{\textbf{Ablation of the final harness on R2R-CE-100}~\cite{qiao2025opennav} (gpt-6-luna and gpt-6.1-sol, default API's reasoning effort, mean $\pm$ s.d. of three runs).
$\Delta$SR: arm minus Seed 0. $\Delta$SR/Cost: $\Delta$SR per dollar per episode. Both: mean of the two LLMs.
M1: visual feedback, M2: start survey, M3: arrival survey, M4: floor check, M5: arrival audit, M6: audit one effort level up.
$\circ$: audit on the current view only.}
\label{tab:abl_arms}
\centering
\footnotesize
\setlength{\tabcolsep}{3pt}
\setlength{\aboverulesep}{0pt}
\setlength{\belowrulesep}{0pt}
\renewcommand{\arraystretch}{1.15}
\resizebox{\textwidth}{!}{%
\begin{tabular}{lcccccccccccccccc}
\specialrule{1pt}{0pt}{0pt}
\rowcolor[gray]{0.9} & \multicolumn{6}{c}{Modules} & \multicolumn{4}{c}{gpt-6-luna} & \multicolumn{4}{c}{gpt-6.1\kern-0.08em-sol} & \multicolumn{2}{c}{Both} \\
\noalign{\vskip-\cmidrulewidth}\cmidrule(lr){2-7}\cmidrule(lr){8-11}\cmidrule(lr){12-15}\cmidrule(lr){16-17}
\rowcolor[gray]{0.9}\multirow{-2}{*}{Arm} & M1 & M2 & M3 & M4 & M5 & M6 & SR$\uparrow$ & SPL$\uparrow$ & $\Delta$SR & $\Delta$SR/Cost$\uparrow$ & SR$\uparrow$ & SPL$\uparrow$ & $\Delta$SR & $\Delta$SR/Cost$\uparrow$ & $\Delta$SR & $\Delta$SR/Cost$\uparrow$ \\
\specialrule{0.5pt}{0pt}{0pt}
\multicolumn{17}{l}{\rqtag{7}\hspace{0.4em}\textit{Which abilities bring the gains?}} \\
Seed 0 & $\times$ & $\times$ & $\times$ & $\times$ & $\times$ & $\times$ & \cellcolor[rgb]{0.984,0.992,1.000}43.7{\scriptsize$\,\pm$5.1} & \cellcolor[rgb]{0.982,0.991,1.000}36.2{\scriptsize$\,\pm$4.7} & -- & -- & \cellcolor[rgb]{0.930,0.964,0.998}72.0{\scriptsize$\,\pm$2.0} & \cellcolor[rgb]{0.947,0.973,0.999}60.5{\scriptsize$\,\pm$2.6} & -- & -- & -- & -- \\
Feedback and tracking only & $\checkmark$ & $\checkmark$ & $\times$ & $\checkmark$ & $\times$ & $\times$ & \cellcolor[rgb]{0.959,0.979,0.999}46.3{\scriptsize$\,\pm$2.5} & \cellcolor[rgb]{0.924,0.961,0.998}39.0{\scriptsize$\,\pm$2.1} & \cellcolor[rgb]{0.999,0.984,0.988}+2.67{\scriptsize$\,\pm$\,3.23} & \cellcolor[rgb]{0.999,0.983,0.988}+183 & \cellcolor[rgb]{0.879,0.938,0.997}74.3{\scriptsize$\,\pm$2.1} & \cellcolor[rgb]{0.864,0.931,0.997}63.8{\scriptsize$\,\pm$2.7} & \cellcolor[rgb]{0.999,0.983,0.988}+2.33{\scriptsize$\,\pm$\,2.77} & \cellcolor[rgb]{0.999,0.981,0.986}+58 & \cellcolor[rgb]{0.999,0.984,0.988}+2.50{\scriptsize$\,\pm$\,2.15} & \cellcolor[rgb]{0.999,0.983,0.987}+92 \\
Arrival verification only & $\times$ & $\times$ & $\checkmark$ & $\times$ & $\checkmark$ & $\checkmark$ & \cellcolor[rgb]{0.982,0.991,1.000}44.3{\scriptsize$\,\pm$11.0} & \cellcolor[rgb]{0.971,0.985,0.999}37.1{\scriptsize$\,\pm$7.8} & \cellcolor[rgb]{0.999,0.984,0.988}+0.67{\scriptsize$\,\pm$\,3.00} & \cellcolor[rgb]{0.999,0.984,0.988}+35 & \cellcolor[rgb]{0.870,0.934,0.997}74.7{\scriptsize$\,\pm$2.1} & \cellcolor[rgb]{0.916,0.957,0.998}62.0{\scriptsize$\,\pm$2.5} & \cellcolor[rgb]{0.999,0.981,0.986}+2.67{\scriptsize$\,\pm$\,2.71} & \cellcolor[rgb]{0.999,0.980,0.986}+59 & \cellcolor[rgb]{0.999,0.984,0.988}+1.67{\scriptsize$\,\pm$\,1.92} & \cellcolor[rgb]{0.999,0.984,0.988}+52 \\
\textbf{NavRSI (final harness)} & $\checkmark$ & $\checkmark$ & $\checkmark$ & $\checkmark$ & $\checkmark$ & $\checkmark$ & \cellcolor[rgb]{0.778,0.886,0.995}\textbf{51.3{\scriptsize$\,\pm$6.7}} & \cellcolor[rgb]{0.778,0.886,0.995}\textbf{41.9{\scriptsize$\,\pm$4.2}} & \cellcolor[rgb]{0.988,0.769,0.832}\textbf{+7.67{\scriptsize$\,\pm$\,3.21}} & \cellcolor[rgb]{0.988,0.769,0.832}\textbf{+426} & \cellcolor[rgb]{0.778,0.886,0.995}\textbf{77.7{\scriptsize$\,\pm$3.5}} & \cellcolor[rgb]{0.778,0.886,0.995}\textbf{66.2{\scriptsize$\,\pm$2.1}} & \cellcolor[rgb]{0.988,0.769,0.832}\textbf{+5.67{\scriptsize$\,\pm$\,2.59}} & \cellcolor[rgb]{0.988,0.769,0.832}\textbf{+124} & \cellcolor[rgb]{0.988,0.769,0.832}\textbf{+6.67{\scriptsize$\,\pm$\,2.19}} & \cellcolor[rgb]{0.988,0.769,0.832}\textbf{+209} \\
\midrule
\multicolumn{17}{l}{\rqtag{8}\hspace{0.4em}\textit{How much does each module add in turn?}} \\
Seed 0 & $\times$ & $\times$ & $\times$ & $\times$ & $\times$ & $\times$ & \cellcolor[rgb]{0.984,0.992,1.000}43.7{\scriptsize$\,\pm$5.1} & \cellcolor[rgb]{0.982,0.991,1.000}36.2{\scriptsize$\,\pm$4.7} & -- & -- & \cellcolor[rgb]{0.930,0.964,0.998}72.0{\scriptsize$\,\pm$2.0} & \cellcolor[rgb]{0.957,0.978,0.999}60.5{\scriptsize$\,\pm$2.6} & -- & -- & -- & -- \\
Add feedback & $\checkmark$ & $\times$ & $\times$ & $\times$ & $\times$ & $\times$ & \cellcolor[rgb]{0.927,0.963,0.998}47.7{\scriptsize$\,\pm$2.1} & \cellcolor[rgb]{0.887,0.942,0.997}39.9{\scriptsize$\,\pm$2.1} & \cellcolor[rgb]{0.999,0.975,0.982}+4.00{\scriptsize$\,\pm$\,2.79} & \cellcolor[rgb]{0.997,0.942,0.957}+284 & \cellcolor[rgb]{0.879,0.938,0.997}74.3{\scriptsize$\,\pm$4.5} & \cellcolor[rgb]{0.926,0.962,0.998}62.2{\scriptsize$\,\pm$3.9} & \cellcolor[rgb]{0.999,0.983,0.988}+2.33{\scriptsize$\,\pm$\,2.03} & \cellcolor[rgb]{0.999,0.979,0.985}+61 & \cellcolor[rgb]{0.999,0.980,0.986}+3.17{\scriptsize$\,\pm$\,1.71} & \cellcolor[rgb]{0.998,0.965,0.974}+121 \\
Add start survey & $\checkmark$ & $\checkmark$ & $\times$ & $\times$ & $\times$ & $\times$ & \cellcolor[rgb]{0.918,0.958,0.998}48.0{\scriptsize$\,\pm$1.0} & \cellcolor[rgb]{0.850,0.923,0.997}40.6{\scriptsize$\,\pm$0.7} & \cellcolor[rgb]{0.998,0.967,0.976}+4.33{\scriptsize$\,\pm$\,3.55} & \cellcolor[rgb]{0.997,0.942,0.958}+283 & \cellcolor[rgb]{0.887,0.942,0.997}74.0{\scriptsize$\,\pm$2.0} & \cellcolor[rgb]{0.913,0.956,0.998}62.8{\scriptsize$\,\pm$1.3} & \cellcolor[rgb]{0.999,0.984,0.988}+2.00{\scriptsize$\,\pm$\,2.63} & \cellcolor[rgb]{0.999,0.983,0.988}+51 & \cellcolor[rgb]{0.999,0.980,0.986}+3.17{\scriptsize$\,\pm$\,2.33} & \cellcolor[rgb]{0.998,0.969,0.977}+117 \\
Add arrival survey & $\checkmark$ & $\checkmark$ & $\checkmark$ & $\times$ & $\times$ & $\times$ & \cellcolor[rgb]{0.970,0.984,0.999}45.7{\scriptsize$\,\pm$3.8} & \cellcolor[rgb]{0.952,0.976,0.999}38.0{\scriptsize$\,\pm$2.9} & \cellcolor[rgb]{0.999,0.984,0.988}+2.00{\scriptsize$\,\pm$\,3.67} & \cellcolor[rgb]{0.999,0.984,0.988}+121 & \cellcolor[rgb]{0.887,0.942,0.997}74.0{\scriptsize$\,\pm$1.7} & \cellcolor[rgb]{0.844,0.920,0.997}65.1{\scriptsize$\,\pm$1.6} & \cellcolor[rgb]{0.999,0.984,0.988}+2.00{\scriptsize$\,\pm$\,2.41} & \cellcolor[rgb]{0.999,0.984,0.988}+47 & \cellcolor[rgb]{0.999,0.984,0.988}+2.00{\scriptsize$\,\pm$\,2.37} & \cellcolor[rgb]{0.999,0.984,0.988}+68 \\
Add floor check (no audit) & $\checkmark$ & $\checkmark$ & $\checkmark$ & $\checkmark$ & $\times$ & $\times$ & \cellcolor[rgb]{0.978,0.989,1.000}45.0{\scriptsize$\,\pm$4.0} & \cellcolor[rgb]{0.929,0.963,0.998}38.8{\scriptsize$\,\pm$1.9} & \cellcolor[rgb]{0.999,0.984,0.988}+1.33{\scriptsize$\,\pm$\,3.10} & \cellcolor[rgb]{0.999,0.984,0.988}+84 & \cellcolor[rgb]{0.903,0.950,0.998}73.3{\scriptsize$\,\pm$3.1} & \cellcolor[rgb]{0.928,0.963,0.998}62.1{\scriptsize$\,\pm$2.0} & \cellcolor[rgb]{0.999,0.984,0.988}+1.33{\scriptsize$\,\pm$\,2.67} & \cellcolor[rgb]{0.999,0.984,0.988}+31 & \cellcolor[rgb]{0.999,0.984,0.988}+1.33{\scriptsize$\,\pm$\,2.02} & \cellcolor[rgb]{0.999,0.984,0.988}+46 \\
Add audit (no lift) & $\checkmark$ & $\checkmark$ & $\checkmark$ & $\checkmark$ & $\checkmark$ & $\times$ & \cellcolor[rgb]{0.896,0.947,0.998}48.7{\scriptsize$\,\pm$3.8} & \cellcolor[rgb]{0.869,0.933,0.997}40.3{\scriptsize$\,\pm$2.7} & \cellcolor[rgb]{0.997,0.946,0.961}+5.00{\scriptsize$\,\pm$\,3.63} & \cellcolor[rgb]{0.996,0.917,0.940}+312 & \cellcolor[rgb]{0.778,0.886,0.995}\textbf{77.7{\scriptsize$\,\pm$1.5}} & \cellcolor[rgb]{0.778,0.886,0.995}\textbf{66.9{\scriptsize$\,\pm$1.5}} & \cellcolor[rgb]{0.988,0.769,0.832}\textbf{+5.67{\scriptsize$\,\pm$\,2.83}} & \cellcolor[rgb]{0.988,0.769,0.832}\textbf{+125} & \cellcolor[rgb]{0.994,0.889,0.919}+5.33{\scriptsize$\,\pm$\,2.32} & \cellcolor[rgb]{0.994,0.875,0.909}+173 \\
\textbf{NavRSI (final harness)} & $\checkmark$ & $\checkmark$ & $\checkmark$ & $\checkmark$ & $\checkmark$ & $\checkmark$ & \cellcolor[rgb]{0.778,0.886,0.995}\textbf{51.3{\scriptsize$\,\pm$6.7}} & \cellcolor[rgb]{0.778,0.886,0.995}\textbf{41.9{\scriptsize$\,\pm$4.2}} & \cellcolor[rgb]{0.988,0.769,0.832}\textbf{+7.67{\scriptsize$\,\pm$\,3.21}} & \cellcolor[rgb]{0.988,0.769,0.832}\textbf{+426} & \cellcolor[rgb]{0.778,0.886,0.995}\textbf{77.7{\scriptsize$\,\pm$3.5}} & \cellcolor[rgb]{0.805,0.900,0.996}66.2{\scriptsize$\,\pm$2.1} & \cellcolor[rgb]{0.988,0.769,0.832}\textbf{+5.67{\scriptsize$\,\pm$\,2.59}} & \cellcolor[rgb]{0.989,0.773,0.834}+124 & \cellcolor[rgb]{0.988,0.769,0.832}\textbf{+6.67{\scriptsize$\,\pm$\,2.19}} & \cellcolor[rgb]{0.988,0.769,0.832}\textbf{+209} \\
\midrule
\multicolumn{17}{l}{\rqtag{9}\hspace{0.4em}\textit{Does removing modules hurt?}} \\
Seed 0 & $\times$ & $\times$ & $\times$ & $\times$ & $\times$ & $\times$ & \cellcolor[rgb]{0.984,0.992,1.000}43.7{\scriptsize$\,\pm$5.1} & \cellcolor[rgb]{0.984,0.992,1.000}36.2{\scriptsize$\,\pm$4.7} & -- & -- & \cellcolor[rgb]{0.964,0.982,0.999}72.0{\scriptsize$\,\pm$2.0} & \cellcolor[rgb]{0.960,0.979,0.999}60.5{\scriptsize$\,\pm$2.6} & -- & -- & -- & -- \\
No start survey & $\checkmark$ & $\times$ & $\checkmark$ & $\checkmark$ & $\checkmark$ & $\checkmark$ & \cellcolor[rgb]{0.918,0.958,0.998}48.0{\scriptsize$\,\pm$7.9} & \cellcolor[rgb]{0.920,0.959,0.998}40.5{\scriptsize$\,\pm$5.8} & \cellcolor[rgb]{0.998,0.967,0.976}+4.33{\scriptsize$\,\pm$\,3.35} & \cellcolor[rgb]{0.998,0.969,0.978}+237 & \cellcolor[rgb]{0.810,0.903,0.996}79.3{\scriptsize$\,\pm$2.1} & \cellcolor[rgb]{0.778,0.886,0.995}\textbf{67.1{\scriptsize$\,\pm$2.1}} & \cellcolor[rgb]{0.992,0.847,0.888}+7.33{\scriptsize$\,\pm$\,2.53} & \cellcolor[rgb]{0.990,0.803,0.857}+172 & \cellcolor[rgb]{0.992,0.850,0.891}+5.83{\scriptsize$\,\pm$\,2.15} & \cellcolor[rgb]{0.991,0.825,0.873}+192 \\
No floor check & $\checkmark$ & $\checkmark$ & $\checkmark$ & $\times$ & $\checkmark$ & $\checkmark$ & \cellcolor[rgb]{0.795,0.895,0.995}51.0{\scriptsize$\,\pm$1.7} & \cellcolor[rgb]{0.903,0.950,0.998}41.0{\scriptsize$\,\pm$0.6} & \cellcolor[rgb]{0.990,0.800,0.854}+7.33{\scriptsize$\,\pm$\,3.40} & \cellcolor[rgb]{0.990,0.803,0.857}+405 & \cellcolor[rgb]{0.920,0.959,0.998}75.0{\scriptsize$\,\pm$4.6} & \cellcolor[rgb]{0.860,0.929,0.997}64.8{\scriptsize$\,\pm$4.2} & \cellcolor[rgb]{0.999,0.984,0.988}+3.00{\scriptsize$\,\pm$\,2.89} & \cellcolor[rgb]{0.999,0.984,0.988}+66 & \cellcolor[rgb]{0.995,0.900,0.927}+5.17{\scriptsize$\,\pm$\,2.20} & \cellcolor[rgb]{0.995,0.900,0.927}+162 \\
No feedback, no floor check & $\times$ & $\checkmark$ & $\checkmark$ & $\times$ & $\checkmark$ & $\checkmark$ & \cellcolor[rgb]{0.828,0.912,0.996}50.3{\scriptsize$\,\pm$2.1} & \cellcolor[rgb]{0.778,0.886,0.995}\textbf{43.4{\scriptsize$\,\pm$0.5}} & \cellcolor[rgb]{0.993,0.852,0.892}+6.67{\scriptsize$\,\pm$\,3.19} & \cellcolor[rgb]{0.993,0.863,0.900}+362 & \cellcolor[rgb]{0.926,0.962,0.998}74.7{\scriptsize$\,\pm$3.1} & \cellcolor[rgb]{0.859,0.928,0.997}64.9{\scriptsize$\,\pm$2.6} & \cellcolor[rgb]{0.999,0.984,0.988}+2.67{\scriptsize$\,\pm$\,3.10} & \cellcolor[rgb]{0.999,0.984,0.988}+58 & \cellcolor[rgb]{0.996,0.930,0.949}+4.67{\scriptsize$\,\pm$\,2.27} & \cellcolor[rgb]{0.997,0.933,0.951}+145 \\
No arrival views & $\checkmark$ & $\checkmark$ & $\circ$ & $\checkmark$ & $\checkmark$ & $\checkmark$ & \cellcolor[rgb]{0.907,0.953,0.998}48.3{\scriptsize$\,\pm$3.5} & \cellcolor[rgb]{0.933,0.966,0.999}40.1{\scriptsize$\,\pm$3.3} & \cellcolor[rgb]{0.998,0.958,0.969}+4.67{\scriptsize$\,\pm$\,3.09} & \cellcolor[rgb]{0.998,0.958,0.969}+259 & \cellcolor[rgb]{0.778,0.886,0.995}\textbf{80.3{\scriptsize$\,\pm$1.1}} & \cellcolor[rgb]{0.798,0.897,0.996}66.6{\scriptsize$\,\pm$1.3} & \cellcolor[rgb]{0.988,0.769,0.832}\textbf{+8.33{\scriptsize$\,\pm$\,3.24}} & \cellcolor[rgb]{0.988,0.769,0.832}\textbf{+181} & \cellcolor[rgb]{0.989,0.787,0.845}+6.50{\scriptsize$\,\pm$\,2.33} & \cellcolor[rgb]{0.989,0.789,0.846}+203 \\
No audit, no lift & $\checkmark$ & $\checkmark$ & $\checkmark$ & $\checkmark$ & $\times$ & $\times$ & \cellcolor[rgb]{0.978,0.989,1.000}45.0{\scriptsize$\,\pm$4.0} & \cellcolor[rgb]{0.966,0.983,0.999}38.8{\scriptsize$\,\pm$1.9} & \cellcolor[rgb]{0.999,0.984,0.988}+1.33{\scriptsize$\,\pm$\,3.10} & \cellcolor[rgb]{0.999,0.984,0.988}+84 & \cellcolor[rgb]{0.948,0.973,0.999}73.3{\scriptsize$\,\pm$3.1} & \cellcolor[rgb]{0.932,0.965,0.998}62.1{\scriptsize$\,\pm$2.0} & \cellcolor[rgb]{0.999,0.984,0.988}+1.33{\scriptsize$\,\pm$\,2.67} & \cellcolor[rgb]{0.999,0.984,0.988}+31 & \cellcolor[rgb]{0.999,0.984,0.988}+1.33{\scriptsize$\,\pm$\,2.02} & \cellcolor[rgb]{0.999,0.984,0.988}+46 \\
No lift & $\checkmark$ & $\checkmark$ & $\checkmark$ & $\checkmark$ & $\checkmark$ & $\times$ & \cellcolor[rgb]{0.896,0.947,0.998}48.7{\scriptsize$\,\pm$3.8} & \cellcolor[rgb]{0.929,0.964,0.998}40.3{\scriptsize$\,\pm$2.7} & \cellcolor[rgb]{0.997,0.946,0.961}+5.00{\scriptsize$\,\pm$\,3.63} & \cellcolor[rgb]{0.996,0.917,0.940}+312 & \cellcolor[rgb]{0.859,0.928,0.997}77.7{\scriptsize$\,\pm$1.5} & \cellcolor[rgb]{0.785,0.890,0.995}66.9{\scriptsize$\,\pm$1.5} & \cellcolor[rgb]{0.997,0.937,0.954}+5.67{\scriptsize$\,\pm$\,2.83} & \cellcolor[rgb]{0.997,0.935,0.952}+125 & \cellcolor[rgb]{0.994,0.889,0.919}+5.33{\scriptsize$\,\pm$\,2.32} & \cellcolor[rgb]{0.994,0.875,0.909}+173 \\
\textbf{NavRSI (final harness)} & $\checkmark$ & $\checkmark$ & $\checkmark$ & $\checkmark$ & $\checkmark$ & $\checkmark$ & \cellcolor[rgb]{0.778,0.886,0.995}\textbf{51.3{\scriptsize$\,\pm$6.7}} & \cellcolor[rgb]{0.865,0.931,0.997}41.9{\scriptsize$\,\pm$4.2} & \cellcolor[rgb]{0.988,0.769,0.832}\textbf{+7.67{\scriptsize$\,\pm$\,3.21}} & \cellcolor[rgb]{0.988,0.769,0.832}\textbf{+426} & \cellcolor[rgb]{0.859,0.928,0.997}77.7{\scriptsize$\,\pm$3.5} & \cellcolor[rgb]{0.812,0.904,0.996}66.2{\scriptsize$\,\pm$2.1} & \cellcolor[rgb]{0.997,0.937,0.954}+5.67{\scriptsize$\,\pm$\,2.59} & \cellcolor[rgb]{0.997,0.936,0.953}+124 & \cellcolor[rgb]{0.988,0.769,0.832}\textbf{+6.67{\scriptsize$\,\pm$\,2.19}} & \cellcolor[rgb]{0.988,0.769,0.832}\textbf{+209} \\
\specialrule{1pt}{0pt}{0pt}
\end{tabular}%
}
\end{table*}

\subsection{Experimental setup}
\label{sec:exp_setup}

\paragraph{Benchmark environments.} Following recent VLN methods~\cite{zhou2026embodied,zhou2026navgpt3}, we evaluate on R2R-CE-100, the 100 val-unseen episodes of R2R-CE~\cite{krantz2020beyond} first sampled by Open-Nav~\cite{qiao2025opennav}, on the full val-unseen splits of R2R-CE (1839 episodes) and RxR-CE~\cite{anderson2020rxr} (3669 English episodes), and on 100 sampled val-unseen episodes of NavRAG-CE~\cite{wang2025navrag}. Following the protocol of agent search and RSI~\cite{hu2024automated,zhang2024aflow,lee2026meta,xia2026rrsi,wang2026adarsi} and the information isolation of \S\ref{sec:isolation}, we use three kinds of sets with separate roles. (1)~The \emph{evolve set} is the search set plus the development set. The \emph{search set} (up to 200 R2R-CE training episodes per round, from training scenes only) is where RSI runs: every candidate is rolled out, diagnosed and scored there, and \textbf{its logs are the only ones the designer reads}. The \emph{development set} (1302 val-unseen episodes) is used to select the final harness among the top candidates, and \textbf{the designer never sees its episodes or how its rewrites score on it}. We \textbf{report no result on the evolve set as held-out}. (2)~The \emph{in-distribution held-out set} is the other 537 val-unseen episodes, including R2R-CE-100, which \textbf{share no path with the development set and enter neither the search nor the selection}. \textbf{All R2R-CE-100 results are thus held-out.} (3)~The \emph{OOD held-out sets} are RxR-CE and NavRAG-CE.

\paragraph{Evaluation metrics.} Following the standard VLN-CE
protocol~\cite{anderson2018vision,anderson2018evaluation,krantz2020beyond,ilharco2019general}, we report navigation
error (NE), oracle success rate (OSR), success rate (SR), success weighted by path length (SPL) and, on RxR-CE,
normalized dynamic time warping (nDTW). Resource use covers model-issued tool calls (Turns), tokens and cost per
episode at list prices.

\paragraph{Implementation details.} RSI uses GPT-6-Luna as $\mathcal{M}_{\ell}$ and GPT-6-Astra (high effort) in Codex CLI as the designer, which writes three candidates per round for eight rounds (\cref{fig:evolution}). The final harness is evaluated with 17 LLMs from four vendors (\cref{tab:r2r_native}), \ie, GPT-5.6-Sol, GPT-6-Luna, GPT-6.1-Terra, GPT-6.1-Sol, GPT-6-Astra, GPT-6.2-Sol and GPT-6.2-Astra from OpenAI, Claude Sonnet-5, Opus-5, Opus-5.5, Fable-5 and Fable-5.2 from Anthropic, Gemini-3.8-Flash and Gemini-4-Hydrogen from Google, and Qwen3.7-Flash, Qwen3.8-Flash and Qwen3.8-27B from Alibaba. Each LLM runs at its default API reasoning effort, except in \cref{tab:r2r_full,tab:navrag,tab:rxr500_eff}, where every
LLM method we run uses GPT-6-Astra at high effort for SOTA comparison (as in NavGPT-3~\cite{zhou2026navgpt3}). The same final harness is used with every LLM and on every benchmark.

\subsection{Comparison with the state of the art}
\label{sec:exp_results}

RQ1--RQ4 compare the final harness (\S\ref{sec:harness}) with prior work.

\rqsec{1}{How does NavRSI compare with SOTA methods?}
\Cref{tab:r2r_full} compares NavRSI with trained models and harness-based methods on the full val-unseen splits.
\textbf{(1)~New SOTA on R2R-CE.} With GPT-6-Astra, NavRSI reaches 85.64 SR and 73.58 SPL, the best among all methods,
and exceeds NavGPT-3 by 4.13 SR (see RQ3 for the in-distribution held-out set). NavGPT-3 pairs the same LLM with an 8B VLA trained on 19.28M samples (6.71M external) and takes
panoramic RGB-D with odometry, while NavRSI uses only a single RGB camera and no training. With the same single RGB input, the data-centric RSI-Nav~\cite{li2026rsinav} retrains a 9B model on self-generated data and reaches 68.6 SR.
\textbf{(2)~Cross-dataset generalization.} The final harness runs unchanged on the OOD held-out sets. It reaches 88.96 SR on RxR-CE
with the best nDTW of 78.97, although most trained methods learn from RxR-CE data. On NavRAG-CE (\cref{tab:navrag}),
NavRSI reaches 39.7 SR with the same LLM as the LLM-based baselines, \ie, +10.3~$\pm$~4.2 over NavGPT-3 and
+5.3~$\pm$~3.2 over MIP (paired, s.e.), while the best trained model reports 31.1 SR on the full set.
\textbf{(3)~Light deployment.} On the same 800 RxR-CE episodes as NavGPT-3's released code (\cref{tab:rxr500_eff}),
NavRSI is on par in SR (+0.5~$\pm$~1.3, paired) but needs 0.93\,GB of GPU memory instead of 18.97\,GB and costs 0.424
instead of 0.562 USD per episode, although NavGPT-3's cost counts its planner only.
Overall, NavRSI makes a frozen LLM
with a single RGB camera a state-of-the-art navigator without any training.

\rqsec{2}{Does the final harness beat native harnesses?}
In \cref{tab:r2r_native}, the same final harness runs unchanged with LLMs from four vendors. \textbf{(1)~Higher SR.} It
outperforms their native harnesses (Codex CLI, Claude Agent SDK, Antigravity and
Qwen Code) by up to 11.3 SR (GPT-5.6-Sol). \textbf{(2)~Lower cost.} It also costs less per episode on most LLMs.
On GPT-6-Luna, GPT-6.1-Terra, GPT-6.1-Sol and GPT-6-Astra, the mean cost drops from 0.2828 to 0.0893 USD
($-$68.4\%) compared with Codex CLI. The final harness is thus LLM-agnostic.

\rqsec{3}{Does NavRSI generalize to held-out sets?}
We examine both kinds of held-out sets defined in \S\ref{sec:exp_setup}.
\textbf{(1)~In-distribution held-out set.} The full split in RQ1 contains the development set used for harness selection. \Cref{tab:r2r_heldout_llms} therefore compares NavRSI with MIP and NavGPT-3 on the other 537 episodes of this split, which were never used in the RSI search or in the selection. All three methods run on the same episodes with the same LLM at its default API's reasoning effort (mean $\pm$ s.d. of three runs), and the two GPT-6.2 LLMs were also never used in RSI. With all three LLMs, NavRSI has the best SR, \eg, 90.3 with GPT-6.2-Astra against 86.8 for MIP and 86.5 for NavGPT-3. Averaged over the three LLMs, its SR is 3.0~$\pm$~0.7 above MIP and 2.8~$\pm$~1.2 above NavGPT-3 (paired over episodes, s.e.). It also uses the fewest tokens per episode with every LLM, \eg, 143k against 514k for MIP and 246k for NavGPT-3 with GPT-6-Astra, while NavGPT-3 adds an 8B VLA with 18.97\,GB of GPU memory.
\textbf{(2)~OOD held-out sets.} On RxR-CE and NavRAG-CE, the final harness runs unchanged with no tuning (RQ1). It is on par with NavGPT-3 on the same 800 RxR-CE episodes (+0.5~$\pm$~1.3 SR, \cref{tab:rxr500_eff}) and above both NavGPT-3 and MIP on NavRAG-CE (+10.3~$\pm$~4.2 and +5.3~$\pm$~3.2 SR, \cref{tab:navrag}).
NavRSI thus generalizes to unseen episodes, datasets and LLMs.

\begin{figure*}[t]
\centering
\vspace{-6pt}%
\includegraphics[width=\textwidth]{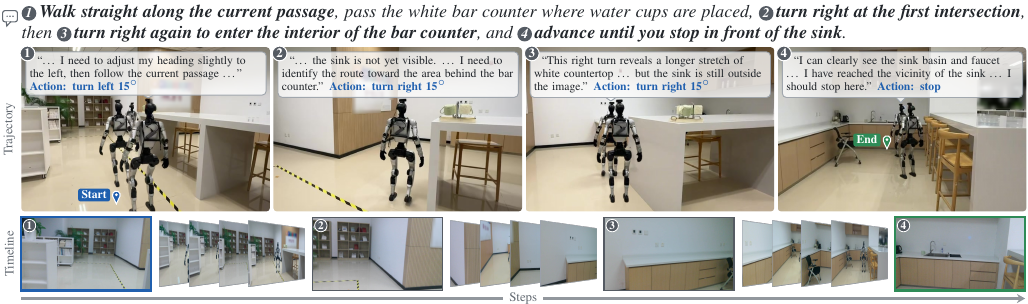}
\caption{{Real-world demo of NavRSI on the Unitree G1 robot.}}
\label{fig:realworld}
\vspace{-12pt}%
\end{figure*}

\rqsec{4}{Does NavRSI transfer to real-world navigation?}
\Cref{tab:realworld} compares NavRSI (single head-mounted RGB camera) with Open-Nav and NavGPT-3 (their own sensors) on a Unitree G1 humanoid robot, over 25 real-world instructions each run four times. NavRSI is the best on all three metrics, reaching 48 SR against 36 for NavGPT-3 and 16 for Open-Nav. \Cref{fig:realworld} shows a typical example, whose long instruction asks the robot to follow a passage,
pass a landmark, turn right twice and stop at a sink. The navigator keeps turning right while ``the sink is still outside the image'', and stops only once ``the
sink basin and faucet'' are in view. The final harness thus works on a real robot without any real-world
data or tuning.

\subsection{Analysis of RSI}
\label{sec:exp_rsi}

RQ5 and RQ6 examine the two parts of RSI: harness rewriting (\S\ref{sec:rsi}) and experience
accumulation (\S\ref{sec:dual}).

\rqsec{5}{Is closed-loop harness rewriting effective?}
\textbf{(1)~Over the seed.} RSI ran with GPT-6-Luna only, yet on every GPT LLM of \cref{tab:r2r_mip}, the final harness
raises SR over Seed~0 by 3.3 (GPT-6.1-Terra) to 9.3 (GPT-5.6-Sol) points with fewer turns. \textbf{(2)~Round by round.}
The best search-set SR rises from 42.5 to 51.5 in three rounds (\cref{fig:evolution}). Stopping RSI earlier also gives a weaker harness on R2R-CE-100 (\cref{tab:abl_rounds}): SR goes from 45.7 at round~0 to 51.3 at round~3 on GPT-6-Luna and from 74.0 to 77.7 on GPT-6.1-Sol, and five more rounds find none better.
\textbf{(3)~Not in one shot.} When the same designer, with the same briefing, writes three harnesses from Seed~0 in one shot, with no closed-loop
feedback (\cref{tab:abl_oneshot}), all three fall below Seed~0 on both LLMs, while NavRSI is +10.89~$\pm$~2.82 SR above
their mean (Holm $p<0.001$). Closed-loop feedback is thus essential.

\rqsec{6}{What does experience accumulation bring?}
We re-run the designer at the start of rounds 4, 6 and 8 and vary only the record it reads
(\cref{tab:abl_exp}). \textbf{(1)~It keeps RSI from repeating itself.} With no record of earlier rounds, 52.8\% of the
designer's new rewrites repeat an earlier one that brought no gain, and by round~8 all of them do. With $E_k$ alone,
none do.
\textbf{(2)~It matches the full record at a fraction of the size.} Here $E_k$ avoids repeated rewrites as effectively as the full record, with 2.2k words in place of 13.1M. \textbf{(3)~It makes the designer cheaper.}
Reading $E_k$ in place of this record also cuts the designer's input tokens per session by 22\% ($p=0.007$). The experience is
thus a compact and sufficient memory of past rewrites, growing by about 340 words per round against 1.5M for the record.

\subsection{Ablation of the final harness}
\label{sec:exp_abl}

RQ7--RQ9 ablate the three abilities and the six modules M1--M6 of the final harness (\S\ref{sec:harness}) on R2R-CE-100 with GPT-6-Luna and GPT-6.1-Sol, and report $\Delta$SR over Seed~0 averaged over both LLMs (\cref{tab:abl_arms}).

\rqsec{7}{Which abilities bring the gains?}
Kept alone, action feedback and progress tracking (M1, M2, M4) add +2.50 SR and arrival verification (M3, M5, M6) adds +1.67, while together they add +6.67, more
than their sum (+4.17). The three abilities are thus complementary.

\rqsec{8}{How much does each module add in turn?}
Adding the modules in the order RSI added them, visual feedback (M1) alone brings +3.17 SR and the start survey (M2) keeps
it there. The arrival survey (M3) and the floor check (M4) lower the gain to +2.00 and +1.33, but the arrival audit (M5) then lifts it to
+5.33, and the audit one effort level up (M6) to +6.67. The arrival audit thus adds the largest gain in the sequence.

\rqsec{9}{Does removing modules hurt?}
Averaged over both LLMs, removing any module from the final harness lowers the gain. The arrival audit matters most: without M5 and M6, the gain falls from +6.67 to +1.33 SR. Without visual feedback and the floor check it falls to +4.67, and without the effort lift (M6) to +5.33.

\section{Conclusion}
\label{sec:conclusion}

We present \textbf{NavRSI}, a harness-centric RSI paradigm for VLN, which recursively
rewrites the navigation harness from its closed-loop logs while accumulating its experience of rewriting. The resulting harness uses only a single RGB camera and primitive actions, and adds no tool, structured memory or agent role. It outperforms the native
harnesses of LLMs from four vendors and, with GPT-6-Astra, sets new SOTA on the full
R2R-CE (85.64 SR) without training, keeps the best SR on held-out episodes, and generalizes to RxR-CE, NavRAG-CE and the real world. Ablations show that its abilities are complementary, that closed-loop RSI outperforms one-shot writing, and that the experience keeps RSI from repeating itself. In future
work, we will extend RSI from the harness to the model itself, so that the two co-evolve from closed-loop experience, and bring it to more embodied tasks beyond navigation.

\clearpage
\makeatletter\global\@colroom\@colht\global\vsize\@colht\makeatother

{
    \small
    \bibliographystyle{ieeenat_fullname}
    \bibliography{main}
}

\end{document}